%% file: 00_main.tex
\pdfoutput=1

\documentclass[11pt]{article}
\PassOptionsToPackage{table,dvipsnames}{xcolor}
\usepackage[final]{acl}

\usepackage{times}
\usepackage{latexsym}

\usepackage[T1]{fontenc}

\usepackage[utf8]{inputenc}

\usepackage{microtype}

\usepackage{inconsolata}

\usepackage{graphicx}

\usepackage{tikz}
\usepackage{listings}
\usepackage{xspace}
\usepackage{booktabs}
\usepackage{wrapfig}
\usepackage{algorithm}
\usepackage{algpseudocode}
\usepackage{subfigure}
\usepackage{amsmath}
\usepackage{enumitem}
\usepackage[many]{tcolorbox}
\usepackage{cleveref}
\crefname{section}{§}{§§}
\Crefname{section}{§}{§§}
\crefformat{section}{§#2#1#3}
\definecolor{lightgray}{gray}{0.9} 

\usepackage{ifthen} 

\newboolean{hidecomments}
\setboolean{hidecomments}{false} 

\newcommand{\hidesentence}[1]{%
  \ifthenelse{\boolean{hidecomments}}{}{#1}%
}

\usepackage{enumitem}
\title{Controllable and Reliable Knowledge-Intensive Task-Oriented\\ Conversational Agents with Declarative Genie Worksheets}

\author{Harshit Joshi  \quad  Shicheng Liu \quad James Chen \quad Robert Weigle \quad Monica S. Lam \\
        Computer Science Department\\
        Stanford University, Stanford, CA \\
        \texttt{\{harshitj, shicheng, lam\}@cs.stanford.edu} \\
{\footnotesize
\includegraphics[width=1em,height=1em]{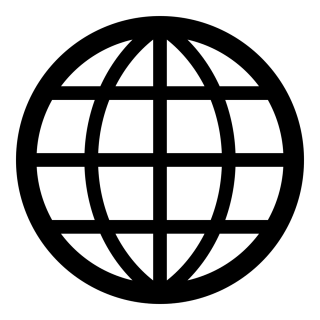}\hspace{0.5em}\url{https://ws.genie.stanford.edu/}} \\
{\footnotesize \includegraphics[width=1em,height=1em]{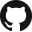}\hspace{0.5em}\url{https://github.com/stanford-oval/genie-worksheets}}
}

\usepackage{newfloat}
\usepackage{caption}

\newenvironment{code}{\captionsetup{type=listing}}{}
\usepackage{setspace}

\newcommand{\system}[0]{Genie\xspace}
\newcommand{\worksheetem}[0]{{\em Genie Worksheet}}
\newcommand{\parserem}[0]{{\em Genie Parser\xspace}}
\newcommand{\runtimeem}[0]{{\em Genie Runtime\xspace}}
\newcommand{\worksheet}[0]{Genie Worksheet}
\newcommand{\parser}[0]{Genie Parser\xspace}
\newcommand{\runtime}[0]{Genie Runtime\xspace}

\newcommand{\RG}[0]{\text{RG}\xspace}
\newcommand{\D}[0]{\mathcal{D}\xspace}
\newcommand{\A}[0]{\mathcal{A}\xspace}
\newcommand{\Report}[0]{\textsc{Report}}
\newcommand{\Confirm}[0]{\textsc{Confirm}}
\newcommand{\Say}[0]{\textsc{Say}}
\newcommand{\Propose}[0]{\textsc{Propose}}
\newcommand{\Ask}[0]{\textsc{Ask}}

\usepackage{xcolor}

\usepackage[colorinlistoftodos]{todonotes}

\usepackage{minitoc}

\definecolor{ourdarkblue}{HTML}{0499CC}
\definecolor{ourlightblue}{HTML}{03A9F4}
\definecolor{ourdarkgray}{HTML}{838A8A}
\definecolor{ourlightgray}{HTML}{B8B8B8}
\definecolor{ourgreen}{HTML}{4D8951}
\definecolor{ourblack}{HTML}{212121}
\definecolor{oursteelblue}{HTML}{9BB8D7}
\definecolor{ourorange}{HTML}{FDBA58}
\definecolor{ourwhite}{HTML}{FAFAFA}
\definecolor{ourpurple}{HTML}{876DB5}
\definecolor{ourmaroon}{HTML}{881C1c}
\definecolor{superlightgray}{HTML}{DDDDDD}

\begin{document}
\doparttoc 
\faketableofcontents

\maketitle

\begin{abstract}
Large Language Models can carry out human-like conversations in diverse settings, responding to user requests for tasks and knowledge. 
However, existing conversational agents implemented with LLMs  often struggle with hallucination, following instructions with conditional logic, and integrating knowledge from different sources.  
These shortcomings compromise the agents' effectiveness, rendering them unsuitable for deployment. 

To address these challenges, we introduce \system{}, a programmable framework for creating knowledge-intensive task-oriented conversational agents.
\system{} can handle involved interactions and answer complex queries.
Unlike LLMs, it delivers reliable, grounded responses through advanced dialogue state management and supports controllable agent policies via its declarative specification -- \worksheet{}. This is achieved through an algorithmic run-time system that implements the developer-supplied policy, limiting LLMs to (1) parse user input using a succinct conversational history, and (2) generate responses according to supplied context.

Agents built with \system{} outperform SOTA methods on complex logic dialogue datasets.
We conducted a user study with 62 participants on three real-life applications: restaurant reservations with Yelp, as well as ticket submission and course enrollment for university students. \system{} agents with GPT-4 Turbo outperformed the GPT-4 Turbo agents with function calling, 
improving goal completion rates from 21.8\% to 82.8\% across three real-world tasks.

\end{abstract}

\input{01_introduction}

\input{02_related_works}

\input{03_architecture}

\input{04_worksheets}

\input{05_agents}

\input{06_static_datasets}

\input{07_evaluation_setup}

\input{08_user_study}

\input{09_conclusion}

\input{10_ethical_considerations}

 \subsubsection*{Acknowledgments}
This work is supported in part by the Verdant Foundation, Microsoft Azure AI credit, Hasso-Plattner Institute, Stanford Human-Centered Artificial Intelligence (HAI) Institute, and JPMorgan Chase.  We also thank Adit Negi, Aryaman Arora, Chenglei Si, Jackie ``Junrui'' Yang, Jiuding Sun, Jordan Juravsky, Junwen Zheng, Ken Liu, Lucia Zheng, Liza Pertseva, Nikil Selvam, Sina Semnani, Tristan Thrush, Yijia Shao, Yanzhe ``Sanju'' Zhang, Yutong Zhang, and the members of the Stanford OVAL (Open Virtual Assistant Lab) for helpful discussion about the project and comments on the manuscript.

\bibliography{custom}

\input{99_appendix}

\end{document}

%% file: 01_introduction.tex
\begin{figure*}[t!]
    \centering
    \includegraphics[width=\textwidth]{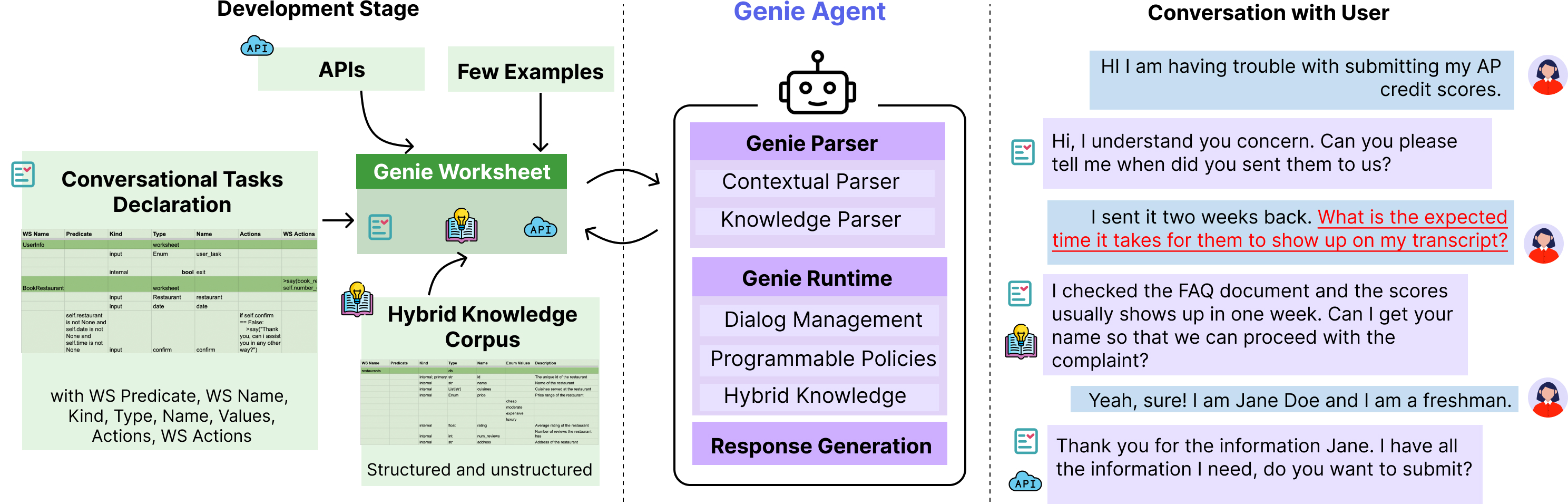}
    \caption{Overview of the Genie Agent. (Left) The developer needs to provide the declarative specification -- \worksheet s with conversational tasks and a knowledge corpus, APIs, and a few examples for the \parser. (Middle) The Genie Parser and Genie Runtime use the \worksheet{} to create a dialogue agent. (Right) The resulting agent can reliably carry out conversations while requiring little developer effort.}
    \label{fig:intro_fig}
    \vspace{-4mm}
\end{figure*}

\section{Introduction}
Large Language Models present a compelling opportunity for building natural, human-like agents. 
Although LLM-based agents can handle ``unhappy paths'' and adeptly respond to unanticipated user inputs at any stage of a conversation, they remain unsuitable for real-world deployment. High-profile failures--such as a Canadian airline being held liable for a chatbot that provided misleading travel advice~\cite{aircandabbc}, or the Cursor agent fabricating policies~\cite{cursor_fortune}--underscore the persistent issue of hallucination and failure to follow predefined policy. 
Recent efforts have been made to mitigate this problem by building knowledge-grounded agents which are capable of querying structured data (e.g., SQL~\citep{pourreza2023dinsql}, SPARQL~\citep{liu2024spinachsparqlbasedinformationnavigation}) and retrieving unstructured text~\citep{khattab2023demonstratesearchpredictcomposingretrievallanguage}. 
Nonetheless, these systems remain constrained to question-answering tasks and lack the capabilities necessary to perform complex, goal-oriented tasks.

Researchers and industry practitioners have created and deployed task-oriented conversational agents.
These agents are typically designed to fill ``slot-values'', such as \texttt{\{restaurant = "Le Bernadin"\}}, based on user utterances to complete a single task~\citep{budzianowski-etal-2018-multiwoz, SMDataflow2020, rastogi2020towards}.
However, such agents cannot handle users' unexpected questions~\cite {rasa2017, converse2022, AmazonLex, BotPress, GoogledialogueFlow}.

We identify three core challenges in deploying reliable and controllable conversational agents.

{\bf Challenge 1: Providing developer control over knowledgeable and responsive agents without onerous efforts.} 
To achieve business objectives, developers desire to maintain control over critical aspects of the agent's operation, including the flow of conversations, the timing of actions, and the information elicited from users.
For example, if a user declines an agent's offer to book a restaurant, developers should be able to program the agent to suggest alternative options, such as offering discounts.
To gain more control over dialogue agents, previous works have added policies as instructions in model prompts~\citep{zhang2023sgp, liu2024llm}.
However, LLMs often fail to adhere strictly to these instructions, resulting in inconsistent and unreliable agent behavior~\citep{liu2024lost}.

{\bf Challenge 2: Support users' knowledge-corpus queries, which may be embedded in a task request.} 
Researchers have created agents for question answering from a knowledge corpus and performing tasks with APIs.
They typically use semantic parsing, which maps natural language to logical forms, for API invocations~\citep{patil2023gorilla} and database queries~\citep{pourreza2023dinsql,
liu2023suql,
gao2023texttosqlempoweredlargelanguage, 
xu2023finetunedllmsknowmore, 
liu2024spinach,
pourreza2024chasesqlmultipathreasoningpreference,
li2024petsqlpromptenhancedtworoundrefinement}.
However, they cannot do both simultaneously.
To integrate knowledge queries and task requests, \citet{kim2020beyond} propose using intent classification to generate a query or an API invocation. 
However, this method cannot support the composition of queries and API calls.
For example, ``I want to book a romantic restaurant in NYC on Valentine's day'' translates into a knowledge query to search for a restaurant, which is then used as an input for a restaurant reservation API. 

{\bf Challenge 3: Must remember pertinent facts from the dialogue history.}
Current LLM-based methods typically use the complete dialogue history ~\citep{ulmer2024bootstrapping, liu2024llm},
or a summary of the conversation~\citep{packer2023memgpt, li2024hello} to contextualize user reponses.
However, complete dialogue histories can cause LLMs to overlook critical details in extended interactions, and summarization may omit essential information.
This can lead to repetitive questioning or generating inaccurate responses, referred to as ``hallucinations"~\citep{bang2023multitask}.

{\bf We propose three key innovations to address the above challenges}.
First, to enable flexible agent behavior -- contrasting with procedural approaches that require explicitly defined logic flows for each scenario -- we introduce a \textbf{novel high-level declarative specification language, \worksheet{}}. This language simplifies the development of reliable task-oriented agents, thereby reducing the burden on developers and enhancing agent adaptability.
The language design allows the composition of knowledge queries and task requests.

Second, our architecture \textbf{integrates LLM-based processes with algorithmic components to achieve conversationality with reliability and control.} The agent's decision-making policy is implemented in an algorithmic run-time system, which is decoupled from the LLM functions. We solely use LLMs to convert natural language utterances into formal representations (semantic parsing) and translate formal directives into naturalistic agent responses (response generation). 
This separation of concerns enables the dialogue agent to remain responsive and naturalistic while adhering to developer-defined rules and ultimately generating grounded responses. 

Third, we have designed a \textbf{{\runtime{}} system that enforces the agent’s policy} by (1) storing critical information and presenting the relevant information for each LLM invocation, (2) running knowledge queries and completing tasks with API invocations, and (3) generating deterministic agent actions that follow developer-defined conversation flow as declared in {\worksheet{}}.

Using \worksheet{}, we have developed \system{} agents that achieve up to a 20.5\% improvement over SOTA methods on StarV2~\citep{zhao2022anytod} across complex logic domains. 
\system{} surpasses LLM function calling from 10.2\% to 40.7\% when using GPT-4 Turbo as the LLM.
Notably, \system{} agents with GPT-4o-mini perform within 3\% of GPT-4 Turbo Genie and outperform GPT-4 Turbo with function calling alone.
We also validate \system{} agents in real-world applications by conducting a user study involving 62 participants across three practical tasks. 
The results show that \textbf{\system{} improves the goal completion rate from 21.8\% to 82.8\% on average.}

%% file: 02_related_works.tex
\begin{figure*}[t!]
    \centering
    \includegraphics[width=\textwidth]{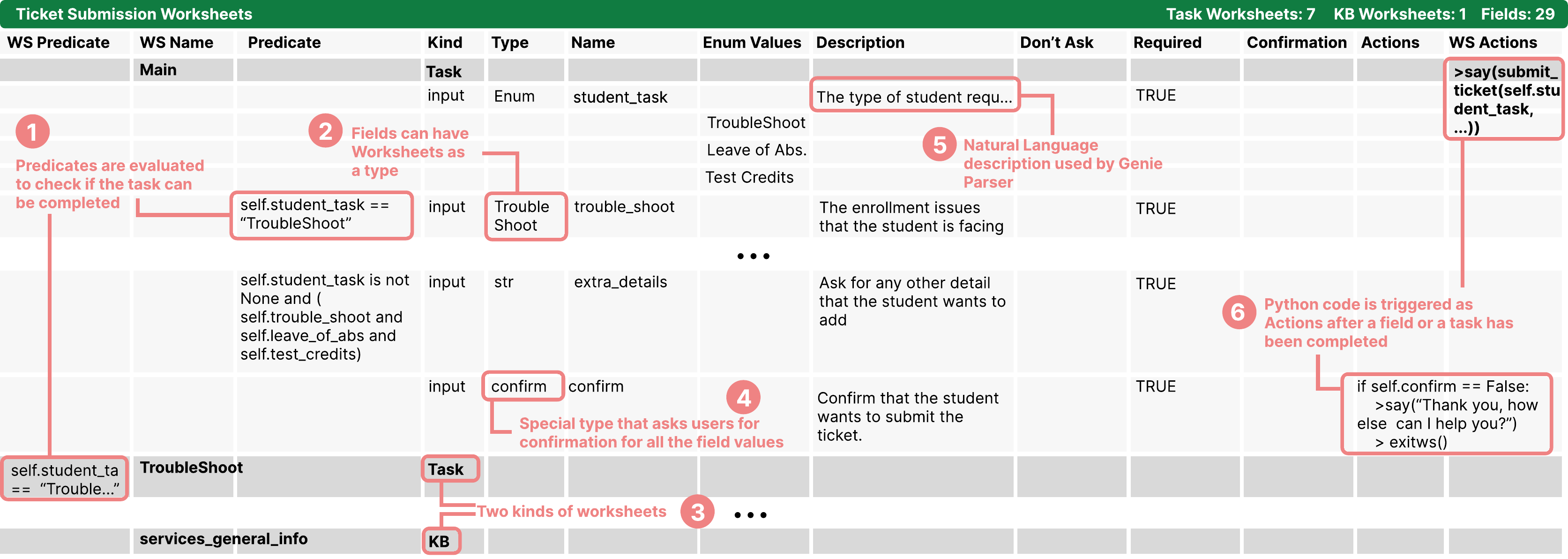}
    \caption{The \worksheet{} specification for the support ticket submission assistant in \autoref{fig:intro_fig}.}
    \label{fig:worksheet}
\end{figure*}

\section{Related Work}

\paragraph{Dialogue Tree-based frameworks} 
A wide variety of commercial and open-sourced products ~\citep{rasa2017, converse2022, MSBot, MSPowerVirtualAssistants, AmazonLex, OneReach, WatsonAssistants, GoogledialogueFlow, BotPress} exist that help developers program a conversational agent using dialogue trees, with slight differences in design. 
For instance, \citet{MSBot,OneReach,WatsonAssistants,BotPress,GoogledialogueFlow} allow users to interactively program with a GUI. 
Developers can manipulate building blocks that classify user intents, sometimes with the help of built-in NLU modules. 
RASA allows developers to declare a list of global intents and program linear conversational paths accordingly ~\citep{rasa2017}. 
Converse allows developers to program a special kind of ``and-or'' trees ~\citep{converse2022}.
In dialogue trees, developers explicitly program all possible dialogue turns and responses, which leads to exponentially many possible dialogue paths and is almost impossible to exhaust in real life.

\paragraph{LLM and task-oriented dialogues}
A series of recent works attempts to integrate LLMs with TOD, showing its capability in dialogue state tracking~\citep{hu2022context, feng2023towards, hudecek-dusek-2023-large, zhang-etal-2023-sgp, li2024large}. In particular, \citet{li2024large} uses GPT's function calling capability for state tracking, which we compare against in Section \ref{sec:eval-with-real-users}. \citet{deng2024plugandplay} uses LLM for its dialogue policy planning module but does not support developer-defined actions through code. When building ToD agents with LLMs, existing works typically feed the entire set of instructions to the model for response generation~\citep{zhang-etal-2023-sgp}. 
While this works well on narrow domains, LLMs struggle to follow all instructions in real-life situations.
\system{} instead deterministically generates formal agent acts such as \texttt{say}, \texttt{report}, \texttt{ask}, as described in Section~\ref{subsec:runtime}, to instruct LLMs on its response.

%% file: 03_architecture.tex
\section{The Genie Architecture}

Genie focuses on 
(1) providing developers with precise control over dialogue agents, 
(2) allowing knowledge access and its composition with task requests, and (3)
handling extended conversation context and following instructions.
It combines four main components, as shown in \autoref{fig:intro_fig}.
(1) \worksheetem{}, a high-level, declarative specification that encapsulates the key elements needed to specify dialogue agents;
(2) \parserem{}, an LLM-based semantic parser that uses the task specification in \worksheet{} to map the user utterances into formal task requests and queries to the knowledge corpus;
(3) \runtimeem{} that converts the specification into formal task definitions and remembers key information in formally defined variables.
It also performs knowledge access using SUQL~\cite{liu2023suql} over structured and unstructured data corpus, and uses information retrieval with vector databases over text documents as needed.
The runtime system forces the agent to follow a predefined policy to generate deterministic agent actions;
(4) Finally, agent acts that are used by the LLM-based \emph{Genie Response Generator} to generate a response.

%% file: 04_worksheets.tex
\section{\worksheet{} Specification}
Conventional dialogue trees require developers to manually implement all possible dialogue paths, rendering such approaches impractical in real-world scenarios~\cite{rasa2017, converse2022, WatsonAssistants, AmazonLex, GoogledialogueFlow}.
To mitigate this limitation, we introduce \worksheet{}, a high-level declarative specification language where developers specify only the information that needs to be supplied by the users and the corresponding actions; there is no need to prescribe the possible dialogue flows. 

There are two kinds of \worksheetem: task worksheets and knowledge worksheets. 
\autoref{fig:worksheet} shows three \worksheetem \emph{s} for a complaint submission portal at a university.
In this example, task worksheets (e.g., \texttt{Main} and \texttt{TroubleShoot}) define the information required from the user, called fields (e.g., \texttt{student\_task} and \texttt{extra\_details}).
In contrast, knowledge worksheets, such as \texttt{services\_general\_info}, declare the knowledge corpus available to the agent for gathering information or filling fields for the user. 
Developers can provide a collection of hybrid knowledge sources by declaring schemas for databases, and supplying a vector database for text. 

Genie Worksheets are analogous to class definitions. As the runtime performs a task or query, it instantiates the corresponding worksheet. The fields in each instance are updated by the \parserem{} and \runtimeem{} to reflect the current dialogue state.
This approach allows the user to supply information in any order they wish, including providing partial information for different API calls and knowledge requests in the same turn.
The agent will track all the information and elicit missing details for each API or knowledge request to be executed.
The \runtimeem{} orchestrates action execution based on the specification's constraints, eliminating the need for developers to draw out all the conversational paths explicitly.

\paragraph{Predicates and Actions}
Each worksheet defines predicates, WS Predicate, that determine when the assistant should gather task-specific information. 
As shown in \autoref{fig:worksheet} (Marker 1), the agent initiates information gathering from the \texttt{TroubleShoot} worksheet when the predicate \texttt{self.student\_task == "Troubleshoot"} evaluates to True. 
This predicate is satisfied when the user indicates they need troubleshooting assistance (captured in \texttt{student\_task}).
\worksheet{} allows predicate-based control on individual fields, enabling fine-grained conditional information gathering.
Once all required field values for a task are collected, the worksheet executes its associated actions (WS Actions) -- such as calling external APIs, updating dialogue state, or unlocking new tasks for the user (see \autoref{fig:worksheet}, Marker 6). 
Additionally, individual fields can have their own actions that execute immediately upon value assignment, providing granular control over the interaction flow.
\autoref{lst:action_example} shows that if the user does not confirm submitting the ticket, then the agent should ask for the reason for not confirming.

\begingroup
\setstretch{0.8}
\begin{code}
\begin{lstlisting}[caption={Action for \texttt{confirm} field similar to \autoref{fig:worksheet}}, label={lst:action_example}]
if self.confirm == False:
    >say("What is the reason?")
    >exitws()
\end{lstlisting}
\end{code}
\endgroup

\paragraph{Field information}
In addition to predicates and actions, developer needs to provide natural language description (Descriptions) of a field and the expected value type (Type) for each field.\footnote{Details on all the definitions are presented in \autoref{appendix:worksheet_desc}} 
They both are provided to the \parserem{} to improve the semantic parsing.
The description contextualizes the fields, and the type definition enforces control over the value stored for each field.

Genie supports several basic data types for field values: \texttt{int}, \texttt{str}, \texttt{date}, \texttt{time}, \texttt{float}, and \texttt{boolean}. 
In addition, it includes a special type, \texttt{confirm}, which behaves similarly to a boolean but requires explicit user confirmation for all field values before executing any associated actions.

Furthermore, to support full composition of tasks and knowledge queries, a field can also be an instance of a worksheet. Its type is the name of the worksheet, and its value is the result to be returned by executing such a worksheet. 
For example, a developer can define a field in a \texttt{RestaurantReservation} task worksheet as ``\texttt{restaurant: Restaurant}'', where \texttt{Restaurant} refers to a knowledge worksheet. 
This implies that the value of the \texttt{restaurant} field should be an instance of the \texttt{Restaurant} knowledge worksheet.

%% file: 05_agents.tex
\begin{figure*}[t]
    \centering
    \includegraphics[width=\textwidth]{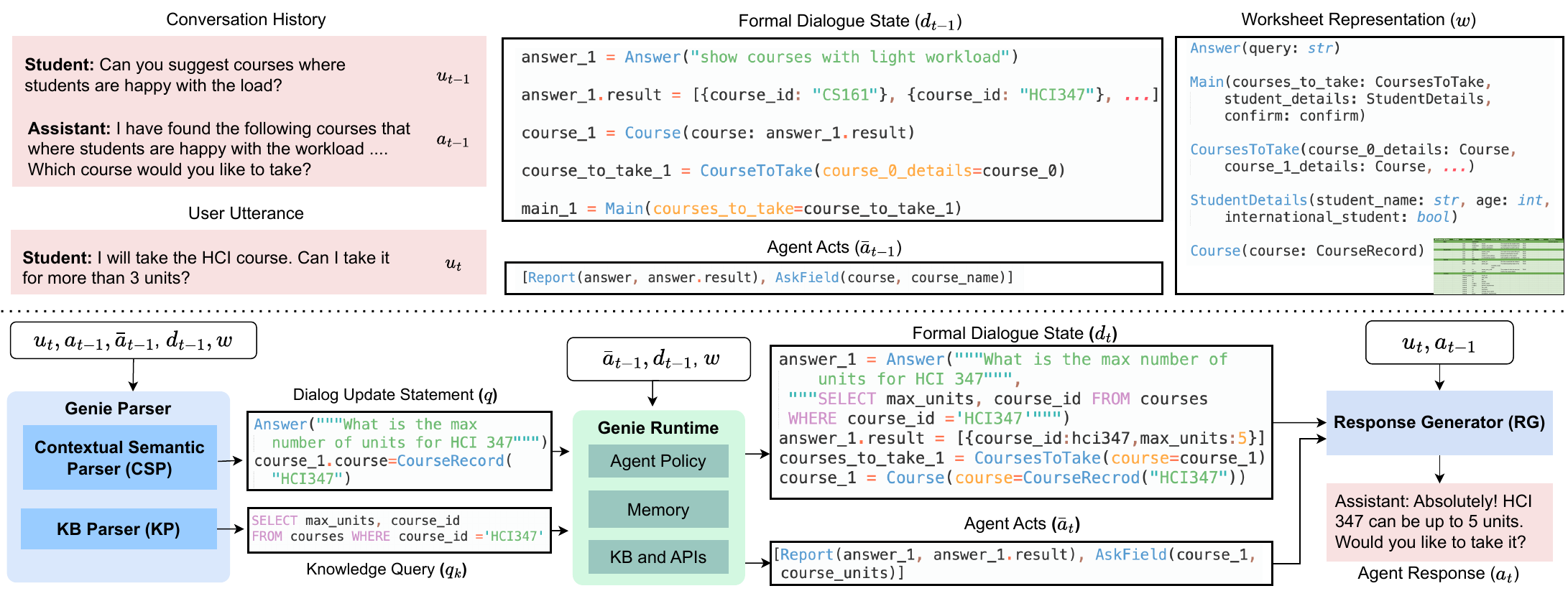}

    \caption{\system{} provides the latest set of worksheets and only one previous conversation turn to the semantic parser. The parser outputs the current worksheets, which are used by the Agent Policy to generate the Agent Dialogue Acts. These Acts along with the latest worksheet value and the latest user utterance are used to generate the response. } 
    \label{fig:methodology}

    \end{figure*}
    
\section{\system{} Dialogue Agents}
We formalize a dialogue as a set of user and agent turns, $\{u_1,a_1, \cdots, u_{t-1}, a_{t-1}, u_{t}\}$, where $u_i \in \mathcal{U}$, $a_i \in \mathcal{A}$ denoting user and agent utterances.

\paragraph{Dialogue State Management}
As discussed as Challenge 3, LLMs struggle with maintaining relevant contextual information across longer conversations. 
To mitigate this issue, \system tracks the dialogue with a formal dialogue state.
As an agent runs, the \parserem{} generates a set of changes to apply to the dialogue state; the \runtimeem{} executes these changes to update the dialogue state, performs actions, and runs knowledge base queries. 
\autoref{fig:methodology} shows the conversation in the course enrollment domain, where the student's goal is to enroll in courses for the next semester.

We represent the dialogue state $D$ as a sequence of states $d_t \in D$ at each turn $t$, where each state contains instances of worksheets performed in that turn. 
Only the instances of the knowledge worksheets referenced in the most recent turn are preserved, and referred to as \texttt{Answer},
while query instances from previous turns are discarded to prevent dialogue state explosion. 
We retain all the task instances.
In our naming convention, the $i$th instance of a worksheet will be referred to as \texttt{<worksheetname>\_}$i$, as shown in \autoref{fig:methodology}\footnote{More details on the representation are in~\autoref{app:naming_convention}.}.

\subsection{\parser{}}
The \parser{} consists of two components.
First, the Contextual Semantic Parser (CSP) translates user utterances into formal dialogue state update statements ($q$) as a sequence of Python statements.
Then, if any knowledge queries exist, the Knowledge Parser (KP) translates the natural language queries into formal knowledge queries ($q_k$).

\begin{figure}[t]
        \centering
        \includegraphics[width=0.9\linewidth]{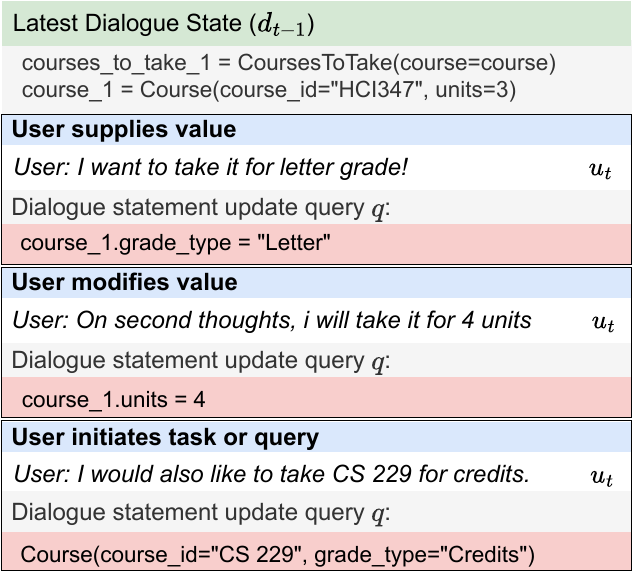}
        \caption{The \parserem{} produces changes to be made to the latest dialogue state $d_{t-1}$, when the user supplies value or modifies them, and creates new instances when the user initiates a new task or query. }
        \label{fig:worksheetquery}
        \vspace{-4mm}
\end{figure}

At any conversation turn, the user may (1) \textit{supply values} to fields that are anticipated in existing worksheets;
(2) \textit{modify} a previously filled field or remove its value;
(3) \textit{initiate} new tasks or queries.
As shown in
\autoref{fig:worksheetquery}, 
if the user provides values for fields or modifies previously filled fields, the queries are update statements that modify the existing state.  
If the user initiates a new task or query, a new statement is created.
The CSP leaves queries to knowledge bases in natural language.

The KP translates any remaining natural language queries into formal knowledge queries such as SUQL.
This function decomposition allows for more advanced methods for knowledge retrieval~\citep{yao2022react,yao2023treethoughtsdeliberateproblem, liu2024spinachsparqlbasedinformationnavigation} and abstracts knowledge integration away from CSP. 
We create a ReAct agent~\cite{yao2022react} and give it access to the knowledge base.\footnote{The implementation details are in \autoref{appendix:experimental_settings}.} 
For example, in \autoref{fig:methodology}, the user's question on the maximum number of units is retained as an NL question in $q$ by CSP. KP then resolves it to executable SUQL (in $d_t$).
The output of this parsing process seamlessly feeds into the \runtimeem{}, where Python statements are executed to get a new dialogue state.

\subsection{\runtime{}}
\label{subsec:runtime}
The \runtimeem{} helps the agent retain all the information mentioned in the conversation and forces the agent to follow the predefined Agent Policy using a predefined set of Agent Acts.

\paragraph{Agent Acts} 
To control the response of the agent, we define a set of formal agent acts ($\bar{a} \in \bar{\mathcal{A}}$):
\begin{itemize}[itemsep=1pt, topsep=1pt, parsep=0pt]
    \item \Report(instance): Report the result obtained by executing the task or query for the instance.
    \item \Confirm(instance, field name): Confirm the value mapped to a field name in the instance. 
        \item \Say(utterance): Explicitly respond with the given utterance.
    \item \Propose(instance): Propose a new task or query to the user, with possibly partially pre-filled values, in the given instance.
        \item \Ask(instance, field name): Ask for the value of a field in an instance.
\end{itemize}
        
\paragraph{Agent Policy} Instead of relying on LLM to directly generate agent responses, \runtimeem{} computes the agent acts to generate agent responses.
The rationale is twofold: 
(1) Providing agent acts causes the LLM to generate a deterministic response governed by the agent policy.
(2) LLMs struggle with under-represented developer-defined policies and cannot follow all the instructions.
\runtimeem{} performs all the developer-provided instructions, such as evaluating predicates, checking types for fields, executing knowledge queries, and executing actions, ensuring that the \system agent can follow all the instructions defined in the \worksheetem{}. 

Once the semantic parsing is performed, the agent policy\footnote{Algorithm~\ref{alg:agent_policy} in \autoref{appendix:agent_policy} outlines the agent policy.} starts 
by assessing whether an \texttt{Answer} query exists in the dialogue state update statements.
If any required parameters are missing in the instance, the agent prompts the user to provide the missing values.
Otherwise, the system reports the result of the \texttt{Answer} query.
Subsequently, the policy checks whether any new fields require confirmation and prompts users.
If no confirmation is needed, actions associated with the newly filled fields are executed,
and the dialogue state is updated accordingly.
If all fields in any worksheet are filled and their corresponding actions have been completed, the policy executes the worksheet's actions 
and updates the dialogue state.
Finally, the policy identifies the first unfilled field in the dialogue state and generates an agent act requesting the user to provide the missing values.

For instance in \autoref{fig:methodology}, the policy first executes the \texttt{Answer} query and adds the result as \Report(\texttt{answer}) act to the agent acts list.
It then checks all the required fields in the \texttt{courses\_to\_take\_1} instance and finds that the value of field ``course\_units'' is unfilled and hence adds \Ask(\texttt{courses\_to\_take\_1}, course\_units).

\subsection{Response Generation}
Once we have the new dialogue state $d_t$ and the agent's dialogue acts $\bar{a_t}$ produced by the agent policy, we generate a response to the user. 
The LLM takes the previous agent utterance and the current user utterance($a_{t-1}, u_t$) as additional context. 
The LLM is instructed to adhere to the agent acts $\bar{a}$ instead of making up actions. 
As shown in \autoref{fig:methodology}, the agent response ($a_t$) ``... HCI 347 can be up to 5 units. Would you like to take it?" corresponds to the \Report{} and \Ask{} agent acts ($\bar{a}_t$) generated by the agent policy.
The prompt is shown in~\autoref{appendix:prompt-for-csp-and-rg}.

\subsection{What Makes \system{} Different?}
\system{} automatically handles the dialogue flow using the succinct declarative \worksheetem{} specification. 
This significantly differs from dialogue trees, which require the programmer to hand-code the policy for each turn.
Another popular approach is to model the conversation as a finite state machine where user inputs are mapped to a small set of predefined user dialogue acts, which are then mapped to agent dialogue acts along with the results of API or DB queries.
Since each turn can invoke multiple agent acts, such a design would result in exponentially many agent dialogue states. 
Additionally, purely function calling in LLMs will fail to retain information in a conversation and handle all the developer instructions (Section~\ref{sec:eval-with-real-users}).\footnote{In~\autoref{sec:challenging} we show some conversational snippets.}

%% file: 06_static_datasets.tex
\section{Evaluation on the StarV2 Benchmark}
\label{sec:eval-on-static-datasets}
To compare with prior work, we compare \system with other baselines on StarV2, the most complex of slot-based Task-oriented dialogue (TOD) benchmarks~\citep{zhao2022anytod}, specifically the
bank, trivia, and trip domains.
In these domains, the agent should perform different tasks based on the user's response to a previous question.
For evaluation, we create one worksheet for each domain and define the APIs in a Python file.

\textbf{Baselines:} We compare ~\system{} to AnyTOD~\citep{zhao2022anytod} the SOTA result on StarV2. 
AnyTOD (AT) finetunes T5 XXL (13B) model on all thirteen domains except the one tested (roughly 6000 data points). (-SGD) refers to additional fine-tuning on Schema Guided Dialog. (-PROG) denotes the use of programmable policies. We also evaluate current SOTA LLMs: Llama 3.1 Instruct 70B, referred to as Llama 3.1 70B in the rest of the paper~\citep{dubey2024llama} and GPT-4 Turbo. We prompt the LLM with a policy described in natural language, the dialogue state as provided in STARv2, agent acts with descriptions, and the user utterance.
We ask the LLM to select the next agent act according to the given input.
We provide experimental details in the Appendix~\ref{app:baseline_settings}. 

\subsection{Results}

From \autoref{table:starv2}, we observe that a zeroshot GPT-4 Turbo-based agent is competitive against a finetuned AnyTOD (AT-SGD XXL) and outperforms it in two domains.
However, AnyTOD with programmable policies outperforms zeroshot agents.
\system{} based on GPT-4 Turbo outperforms all the baselines across the domains tested, beating the previous SOTA by 17.5\%, 20.5\%, and 6.4\% in the bank, trip, and trivia domains, respectively.
This validates that \system{} subsumes the slot-filling paradigm.
Furthermore, our analysis indicates that most of our errors are caused by inconsistent data annotations (\autoref{app:errors_starv2}), also highlighted by~\citet{zhao2022anytod}.

\subsection{Improvement over Base LLMs} In ~\autoref{table:starv2}, we also show how different base LLMs affect the performance of \system{} using Llama 3.1 70B, GPT-4o-mini and GPT-4 Turbo.
We observe that \system{} significantly improves the performance of all the base models across all domains by from 0.5 to 33.2 for Llama 3.1 70B, 20.5 to 36.7 for GPT-4o-mini and 10.2 to 40.7 points for GPT-4 Turbo, respectively. 
The experiments also highlight that weaker models like Llama 3.1 70B and GPT-4o-mini can perform similarly to much stronger models like GPT-4 Turbo in some domains. 
For example, in the banking domain, \system{} with GPT-4o-mini matches the performance of GPT-4 Turbo and Llama 3.1 falls short by only 0.4 points.
However, in the Trip and Trivia domains, we observe a bigger gap.
Additionally, using \system{} with GPT-4o-mini beats using only GPT-4 Turbo in all domains.

\begin{table}[t]
    \small
    \centering
    \begin{tabular}{@{}lrrr@{}}
    \toprule
    \textbf{Agent}   & \textbf{Bank} & \textbf{Trip} & \textbf{Trivia} \\
    \rowcolor{lightgray}
    \multicolumn{4}{c}{\textit{Finetuned T5 (11B)}} \\
    AT XXL           & 54.3          & 52.4          & 73.8            \\
    AT-SGD XXL       & 53.1          & 51.5          & 81.1            \\
    AT-PROG XXL      & 61.0            & 60.8          & 73.7            \\
    AT-PROG +SGD XXL & 65.0            & 62.9          & 86.3            \\ 
    \rowcolor{lightgray}
    \multicolumn{4}{c}{\textit{Zeroshot}} \\
    Llama 3.1 70B (FC) & 48.9 & 41.7 & 81.7 \\ 
    GPT-4o-mini (FC) & 50.8 & 43.8 &	69.8 \\
    GPT-4 Turbo (FC) & 55.1 & 42.7 & 82.5 \\ 
    \midrule
    \system{} + Llama 3.1 70B \textbf{(Ours)}         & 82.1         & 75.9         & 82.2           \\
    \system{} + GPT-4o-mini \textbf{(Ours)} & \textbf{82.5} &	\underline{80.5} &	\underline{90.3} \\
    \system{} + GPT-4 Turbo \textbf{(Ours)}          & \textbf{82.5}         & \textbf{83.4}         & \textbf{92.7}           \\ \bottomrule
    \end{tabular}
    \caption{System Action F1 for complex logic domains in StarV2. \system{} improves base LLMs, with Genie + GPT-4o-mini beating GPT-4 (FC) in all three domains and Genie + Llama beating GPT-4 (FC) in two domains. {\bf Bold} is best and \underline{underline} is second best.}
    \label{table:starv2}
\end{table}

%% file: 07_evaluation_setup.tex
\begin{table*}[t!]
\small
\centering
\begin{tabular}{@{}lrr|rr|rr|rr@{}}
\toprule
\textbf{} & \multicolumn{2}{c}{\textbf{All Domains}}                                     & \multicolumn{2}{c}{\textbf{Restaurant}}                                      & \multicolumn{2}{c}{\textbf{Ticket Submission}}                               & \multicolumn{2}{c}{\textbf{Course Enrollment}}                               \\ \midrule
          & \multicolumn{1}{c}{\textbf{GPT4 (FC)}} & \multicolumn{1}{c}{\textbf{Genie}} & \multicolumn{1}{c}{\textbf{GPT4 (FC)}} & \multicolumn{1}{c}{\textbf{Genie}} & \multicolumn{1}{c}{\textbf{GPT4 (FC)}} & \multicolumn{1}{c}{\textbf{Genie}} & \multicolumn{1}{c}{\textbf{GPT4 (FC)}} & \multicolumn{1}{c}{\textbf{Genie}} \\ \cmidrule(l){2-9} 
S. Parsing Accuracy    & \multicolumn{1}{r}{-}                   & 91.4                               & \multicolumn{1}{r}{-}                   & 93.8                               & \multicolumn{1}{r}{-}                   & 85.8                               & \multicolumn{1}{r}{-}                   & 94.1                               \\
Execution Accuracy    & 65.4                                    & \underline{{\bf 86.5}}                               & 50.0                                      & \underline{{\bf 88.0}}                                 & 58.6                                    & \underline{{\bf 80.0}}                                 & 79.2                                    & \underline{{\bf 89.7}}                               \\
Dialouge Act Accuracy    & 69.1                                    & \underline{{\bf 89.2}}                               & 57.7                                    & \underline{{\bf 92.5}}                               & 67.6                                    & \underline{{\bf 82.5}}                               & 77.8                                    & \underline{{\bf 92.9}}                               \\
Goal Completion Rate   & 21.8                                    & \underline{{\bf 82.8}}                               & 54.5                                    & \underline{91.6}                               & 0.0                                       & \underline{{\bf 80.0}}                                 & 10.0                                      & \underline{{\bf 80.0}}                                 \\ \bottomrule
\end{tabular}
\caption{We perform a real user study with 62 users across three domains. We find that \system{} performs significantly better than GPT-4 turbo with Function Calling ability on the three applicable metrics. We perform t-test and mark the results with $p < 0.05$ with {\bf Bold}. \underline{Underline} represents the best result.  }
\label{table:main}

\end{table*}

\section{Evaluation on Real User Studies}
\label{sec:eval-with-real-users}

\subsection{Real-World Applications}
Since StarV2 is an artificial benchmark, we study three diverse real-world applications that are significantly more complex, and cannot be handled by slot-based agents. (See Table~\ref{table:stats} in the Appendix.) The evaluation is performed with real humans. 

\noindent\textbf{Restaurant Reservation}
Making a restaurant reservation requires finding a suitable restaurant and providing booking details to complete a transaction. Here, users are expected to talk like they are dealing with human agents, and are not limited to ask about or provide values for prescribed slots. 
We use the real-life dataset containing restaurants from Yelp.com from ~\citet{liu2023suql}. 
  
\noindent\textbf{Ticket Submission}
In this study, we aim to replicate a subset of tasks found within a university's student services portal. 
These portals contain various tasks and subtasks.
Moreover, they often contain a corpus of free-text data, which students must peruse before submitting a ticket.
We evaluate agents' capability to handle multiple tasks.

\noindent\textbf{Course Enrollment}
We evaluate the performance of \system{} as a course enrollment assistant, which combines hybrid data sources to search for course details and fill out complicated nested forms.
The assistants allow students to ask questions about course requirements, student reviews, and ratings while filling out their enrollment forms.
We collect a real-life dataset containing courses from the Computer Science program, with four tables.

\subsection{Experimental Setting}

\noindent\textbf{Baselines:} We compare our system against GPT-4 Turbo with functional calling, which we call GPT-4 (FC).
We provide the baseline system with the ability to use the same Knowledge Parser as \system{} and to execute SUQL for external knowledge access over a hybrid structured and unstructured data corpus.
This baseline closely follows ~\citet{li2024large}. We also experimented with Nemo-Guardrails~\citep{rebedea-etal-2023-nemo}, a framework for LLM-based conversational agents, but found it fails miserably on basic task-oriented dialogues (refer to \autoref{sec:details-on-nemo} for details).

\noindent\textbf{Study Design:} We use Prolific to recruit 42 participants for Restaurant Reservation and Ticket Submission. 
We recruited 20 university students to evaluate the Course Enrollment assistant. 
We instructed participants to book a reservation, submit an issue ticket, and enroll in two courses, respectively.
We randomly assigned users to one of the two visually identical systems and collected 99, 81, and 127 turns with \system{}, and 90, 70, and 144 turns with GPT 4 (FC), respectively.

\noindent\textbf{Evaluation Metrics:} We use four evaluation metrics and manually inspect the conversations to assess each metric. (1) Semantic Parsing Accuracy (SP) is used to evaluate if the natural language user utterance corresponds to the correct APIs, databases, and filled fields. (2) Execution Accuracy (Ex) to check whether the agent executes the correct API and databases. (3) Agent Dialogue Act Accuracy (DA) checks if the agent follows the provided policies. (4) Goal Completion Rate (Goal CR) to evaluate the user’s ability to successfully complete the task with the system’s assistance.

%% file: 08_user_study.tex
\subsection{Main Results}
\label{sec:main_results}
We compare \system{} against GPT 4 (FC) and find that \system{} performs significantly better than GPT-4 (FC) on all three domains, as shown in Table~\ref{table:main}.
We observe that \system{} consistently demonstrates a high semantic parsing rate (exceeding $85\%$), indicating that \worksheet's representation is simple to understand for LLMs with few-shot examples.
The marginally lower Semantic Parsing Accuracy  ($85.8\%$) observed in the Ticket Submission application can be attributed to the application's complexity, featuring several worksheets, as shown in Table~\ref{table:stats} in the \autoref{app:diff_starv2}. 
Since GPT-4 Turbo (FC) responds to the user directly, we do not report the Semantic Parsing Accuracy.

\begin{figure}[t!]
\centering 
        \includegraphics[width=\linewidth, height=4cm]{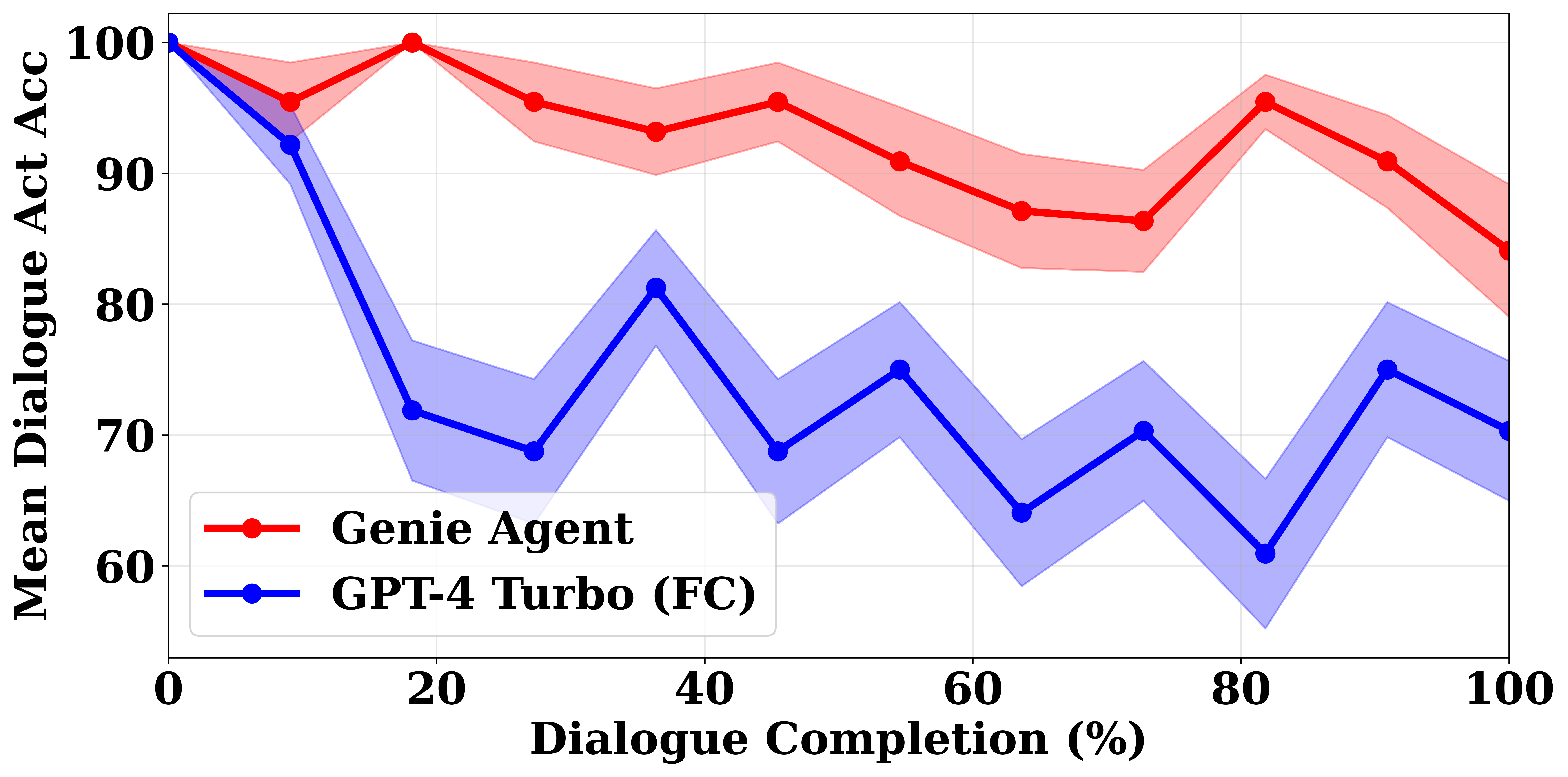}
        \caption{The mean Dialogue Act Acc for \system{} and GPT-4 (FC) as a function of the conversation turns. Genie agents' accuracy is constantly over 80\%.
        }
        \label{fig:conv_analysis}
        \vspace{-2mm}
\end{figure}

Additionally, the higher Execution Accuracy of 86.5\% can be attributed to providing compressed context as the formal dialogue state, enabling the LLM to invoke the correct API and execute the suitable knowledge query.
The superior performance on Dialogue Act Accuracy can be ascribed to our agent's capability to provide turn-by-turn instructions rather than presenting all the instructions at once, as is the case with GPT-4 (FC). 
\textbf{The results validate the benefit of programmable policies in delivering a reliable agent}.

Genie Agents gets a goal completion rate of 82.8\% and beats GPT-4 (FC) with a goal completion rate of 21.8\%.
We observe that GPT-4 (FC) scores relatively higher (54.5\%) in Restaurant Reservation than other domains, the domain it is most familiar with and well represented in existing dialogue datasets~\citep{ye-etal-2022-multiwoz,rastogi2020towards}.
Similar results are also observed by \citet{zhang2023sgp} and \citet{hudecek-dusek-2023-large}. 
However, GPT 4 (FC) struggles with less familiar domains, like Ticket Submission and Course Enrollment.
We hypothesize that the higher number of predicates, which require following several instructions in Ticket Submission, makes it highly challenging for GPT-4 (FC) to assist users.

\begin{figure}[t!]
\centering
        \includegraphics[width=\linewidth]{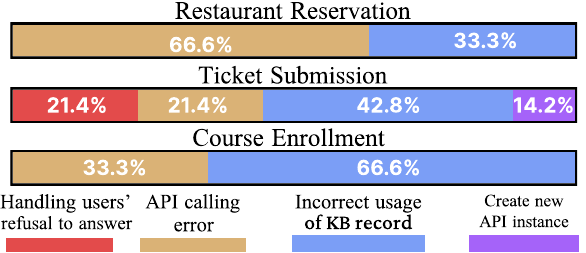}
        \caption{Analysis of semantic parsing errors with \system{} for all the agents. Most of the errors occur due to the mishandling of knowledge queries.
}
        \label{fig:error_analysis}
        \vspace{-2mm}
\end{figure}

\subsection{Error Analysis}
We analyze the mean Dialogue Act Accuracy as a function of the fraction of turns in the conversation for both \system{} and GPT-4 (FC) in Figure~\ref{fig:conv_analysis}.
We observe that \system{} experiences a comparatively smaller performance drop, approximately 8\%, compared to GPT-4’s 43\%. 
This supports our hypothesis that LLMs struggle to maintain context over long interactions, while employing a formal dialogue state mitigates some of these issues. 
We provide examples in the \autoref{app:examples}.

{\bf GPT-4 (FC)}, despite having access to the Knowledge Parser and the SUQL for knowledge access, \textbf{often hallucinates} non-existent information and fails to complete the tasks. We note that it hallucinates in (7/10), (4/11), and (5/10) conversations for course enrollment, ticket submission, and restaurant reservation domains.
Moreover, it prematurely invokes the submission API in all three domains and disregards instructions, such as failing to seek confirmation before invoking APIs.

{\bf \system{} agent} demonstrates high reliability, with overall semantic parsing error rates ranging from just 6.2\% to 14.2\% across different applications.
From~\autoref{fig:error_analysis}, focusing on the Ticket Submission, we observe that the most common failure (21\%) occurs when the agent is unable to proceed because the user refuses to provide required information. Another 21\% of errors in this setting are attributed to incorrect API calling.
The largest source of error (43\%) is due to incorrect generation or misuse of the \texttt{Answer} action, such as querying irrelevant knowledge sources or omitting queries.
Additionally, in 14\% of cases, the parser failed to start a new worksheet, instead updating an existing one incorrectly.
For the Course Enrollment and Restaurant Reservation agents, errors are primarily split between incorrect API calls and improper use of the \texttt{Answer} action: 33\% and 66\% for Course Enrollment and 66\% and 33\% for Restaurant Reservation.

\subsection{Semantic Parsing Error Handling}

To handle errors in semantic parsing, the agent responds to a query by explicitly referencing the interpreted question, enabling users to detect discrepancies.
Similarly, all actions with side effects are explicitly confirmed before execution.
This design allows users to provide corrective input in subsequent turns, prompting the agent to update the dialogue state accordingly.
Even incorrect entries in the dialogue state can facilitate disambiguation and error correction by the semantic parser and should not be discarded prematurely.
We provide examples in the~\autoref{appendix:error_handling_examples}.

%% file: 09_conclusion.tex
\section{Conclusion}

\system{} enables the creation of knowledge-intensive and task-oriented conversational agents that effectively handle complex interactions and queries. By leveraging advanced dialogue state management and a declarative specification, \system{} ensures reliable, grounded responses with controllable agent policies. Our experiments demonstrate that \system{} outperforms SOTA methods on complex logic dialogue datasets. A user study with 62 participants showed significant improvements over GPT-4 Turbo with function calling in real-world tasks.
Unlike LLMs, Genie delivers reliable, grounded responses, and as opposed to traditional dialogue tree agents, it is robust and helpful.

%% file: 10_ethical_considerations.tex
\section*{Limitations}

As an LLM-based system, \system{}'s performance is directly tied to the capability of the underlying LLM. The non-deterministic nature of Language Models can also lead to performance variations across different runs.

Genie scales well to large knowledge bases. For example, the two databases we use for the restaurant and course enrollment domains contain 14 and 35 columns with 1828 and 4214 rows, respectively. However, we observed a slight performance drop as we increase the number of worksheets from 2 (restaurant) to 7 (ticket submission). 
We leave methods to scale Genie to even more complex tasks for future research.

Being a programmable framework, the performance of \system{} is influenced by the specific policies implemented by developers for new domains. The selection of few-shot examples is also expected to impact the agent's performance. To address this, we provide examples of different agents created using our GenieWorksheets in our repository. 

We acknowledge that the additional intermediate LLM calls introduced by \system{} will lead to higher costs and increased latency. However, we believe these trade-offs are necessary to deliver a reliable and accurate integrated task and knowledge assistant.

\section*{Ethical Considerations}

Large Language Models have been increasingly utilized to build various task-oriented agents. We propose a novel method to enhance their accuracy and reliability. We do not anticipate any harm resulting from this work.

For the user evaluation of the Course Enrollment application, we recruited university students who voluntarily participated in the study, awarding each participant an Amazon gift card worth \$10 per 15 minutes. For the Restaurant Reservation and Ticket Submission applications, we used the Prolific platform to recruit participants and compensated them fairly, beyond the minimum wage. Our procedure has been approved by the Institutional Review Board (IRB) at our institution. All collected information is anonymous. We also remove any personal identifiable information from the collected dialogues.

Our code will be released publicly under the Apache License, Version 2.0, and the collected data will be made available to the community.

We use Cursor/Co-pilot for code suggestion while implementing \system.

%% file: 99_appendix.tex
\onecolumn
\newpage
\appendix
\renewcommand \thepart{}
\renewcommand \partname{}
\noptcrule
\part{Appendix} 
\parttoc 
\newpage

\section{Agent Policy Algorithm}
We formally define our Agent Policy in Algorithm~\ref{alg:agent_policy}.
\system{} uses this predefined agent policy and focuses on prioritizing user responses (Line 4) and after resolving user responses it takes actions such as asking next question or calling APIs.

\label{appendix:agent_policy}

        \begin{algorithm}[h]
        \small
        \caption{\system{} Agent Policy}
        \label{alg:agent_policy}
        \begin{algorithmic}[1]
        
        \State \textbf{Input:} Previous dialog state $d$, dialogue state update statements $q$, worksheet: $w$
        \State \textbf{Output: } New dialog state $d'$, Agent Act $\bar{a}$
        \State Initialize a new Agent Acts set $\bar{a} \gets \text{empty set}$
        \If {\texttt{Answer} instance $r$ exists in $q$}
            \If {required parameter $p$ associated with $r$ is unfilled}
                \State $\bar{a} \gets \bar{a} \cup \{\Ask(r, p)\}$ \Comment{Ask for the first required parameter}
            \Else
            \State Execute the \texttt{Answer} query in the \texttt{Answer} instance $r$
            \State $\bar{a} \gets \bar{a} \cup \{\Report(r)\}$ \Comment{Report the result of KB record to the user}
            \EndIf
        \EndIf
            \ForAll {assigned fields $f$ in worksheets present in  $q$ }
                \If {$f$ requires confirmation}
                    \State $\bar{a} \gets \bar{a} \cup \{$\Confirm$(r, f)\}$ \Comment{Ask for confirmation}
                \Else
                    \State Execute the actions associated with the field $f$
                    \State Update the state $d'$ and add new agent acts to $\bar{a}$
                \EndIf
            \EndFor

            \ForAll {Task records $r$ that are complete but whose actions have not been executed}
                 \State Execute the actions associated with the worksheet for $r$
                    \State Update the state $d'$ and add new agent acts to $\bar{a}$
            \EndFor

            \ForAll {required fields $f$ in the records $r$ in  $d'$}
            \If {$f$ has not been assigned}
                \State $\bar{a} \gets \bar{a} \cup \{\Ask(r, f)\}$ \textbf{break} \Comment{Ask for the first unfilled field}
            \EndIf
            \EndFor
        \State \textbf{Return} updated dialog state $d'$ and $\bar{a}$        \end{algorithmic}
        \end{algorithm}

\section{Worksheet Description}
\label{appendix:worksheet_desc}

A worksheet within \worksheet{} can be of two types: task worksheet or knowledge worksheet. The \textbf{Kind} column declares what kind of worksheet is being defined. The \textbf{WS Name} defines the name fo the worksheet, for example ``Main'', ``TroubleShoot'', and ``services\_general\_info'', in~\autoref{fig:worksheet}.
Finally, the action to execute once all the fields are filled and the user has confirmed the values (if required), are declared under \textbf{WS Actions}

For each task worksheet, a field has the following attributes:
\begin{itemize}[itemsep=1pt, topsep=1pt, parsep=0pt]
\item \textbf{Predicate}: Python code that is evaluated to check if the value for this field can be filled in.
\item \textbf{Kind}: \system{} allows three kinds of fields. 
The user provides values for \texttt{input} fields; the \texttt{output} fields are set according to the output of executing an API or knowledge base call; and the \texttt{internal} fields can only be manipulated by the agent. 
\item \textbf{Description} field provides a natural language description of the field, this is used by the \parserem{}.
\item \textbf{Type}: Each field is tied to a type, where all conventional types are allowed. 
\item \textbf{Enum Values}: If it is an enum {\em Type} (e.g., student\_task in \autoref{fig:worksheet}), a set of enumeration values are specified in the succeeding lines. 
\item \textbf{Dont Ask}: A boolean that records the information if the user offers it, but the agent will not ask for it. The "Don't Ask" attribute is crucial for capturing user-provided information that the assistant does not explicitly request but should note if mentioned. For instance, a restaurant reservation assistant might not proactively inquire about seating preferences but would retain such details if the user provides them spontaneously during the conversation. 
\item \textbf{Required}: A boolean which if False, the agent will ask for the value but will allow the user to proceed without answering it. However, if it is set to True, the agent will not proceed without the value for the field.
\item \textbf{Confirmation}: If set to True, the agent asks for confirmation of the value once it set a value. 
\item \textbf{Action}: Finally, whenever a field is assigned, similar to {\em WS Actions}, the supplied {\em Actions} should be executed either if the {\em Confirmation} boolean is set to false or if the user has confirmed with the agent. 
\end{itemize}

For both the worksheet action and field action, the developer can write code using the variables defined in the worksheet (in Python). Several built-in actions are provided to the developer: 
(1) {\tt say (str)} responds to the user with the given string {\tt str}. 
(2) {\tt propose (ws, [fld,val]*)} instantiates a new worksheet {\tt ws} with the given field value pairs. For instance, developers can code in a \texttt{propose} action in the {\em WS Actions} field of a flight booking worksheet to propose a hotel booking worksheet at the same destination of the flight.
(3) {\tt exitws()} marks the worksheet as abandoned and closes it. For instance, the {\em Actions} field of the ``confirm'' field in the ``Main'' worksheet in \autoref{fig:worksheet} contains a {\tt exitws()} call if user does not want to submit the ticket anymore. 

\section{Creating a Genie Agent}
To create a Genie Agent, the developer needs to provide the following:
\begin{itemize}
    \item Genie Worksheets: A specification of all the tasks and knowledge corpus the agent has access to. The agent will follow all the instructions provided here.
    \item API: The APIs declaration are provided in a python file. 
    \item Knowledge Corpus: For knowledge corpus, the developer needs to provide the configuration such as connection uri, or the endpoint of the vector database
    \item Prompts: The developer also needs to provide few-shot examples and guidelines for the semantic parser and the response generator. 
\end{itemize}

\section{Dialogue State naming convention}
\label{app:naming_convention}
Task worksheets are assigned to variables following a systematic naming convention: the worksheet's camel-case name is converted to snake-case and concatenated with an incrementing instance counter.
For example, the first record of \texttt{CoursesToTake} is assigned to \texttt{courses\_to\_take\_1}, with subsequent instances incrementing the counter (e.g., \texttt{courses\_to\_take\_2}). 
Knowledge Base worksheets are represented as \texttt{Answer} and share a common variable name \texttt{answer}.
The dialog state additionally maintains \texttt{answer.result}, which stores the query output from the most recent Knowledge Base interaction.

\section{Formal Representation of \system{}}
The Genie agent has three components: 
\begin{itemize}
\item  $\text{P}(u_t, a_{t-1}, \bar{a}_{t-1}, d_{t-1}, w): \mathcal{U} \times \mathcal{A} \times \mathcal{P}(\bar{\mathcal{A}}) \times \mathcal{D} \times \mathcal{W} \rightarrow \mathcal{D}$: The \parserem{} $P$ 
 accepts the current user utterance $u_t$ and contextualizes the utterance using the last agent utterance $a_{t-1}$, the previous-turn agent acts $\bar{a}_{t-1}$, the formal dialogue state from the last turn $d_{t-1}$, worksheet $w$, and generate a new dialogue state $d$.

\item AP$(d, d', w): \mathcal{D} \times \mathcal{D} \times \mathcal{W} \rightarrow \mathcal{P}(\bar{\mathcal{A}}) \times \D$: an agent policy that accepts the dialogue state from the previous turn $d$, the new dialogue state after parsing $d'$, and the worksheet $w$ and produces a set of agent dialogue acts $\bar{a}$, along with a new dialogue state. 

\item $\RG(d_t, \bar{a}_t, u_t, a_{t-1}): \D \times \mathcal{P}(\bar{\A}) \times \mathcal{U} \times \A \rightarrow \A$: a response generator that accepts the formal dialogue state
$d_t$, an agent act set $\bar{a}_t$, the last user $u_t$ utterance and agent utterance $a_{t-1}$ to generate the agent response $a_t$.
\end{itemize}

Figure \ref{fig:methodology} shows the overview the \system system. At turn $t$ in the conversation, the system is given: (1) the conversation history $\{u_1,a_1, \cdots, u_{t-1}, a_{t-1}, u_{t}\}$, where $u_t \in \mathcal{U}$, $a_t \in \mathcal{A}$ denoting user and agent utterances; (2) the $t-1$-th turn formal agent act $\bar{a}_{t-1} \in \mathcal{P}(\bar{\mathcal{A}})$; (3) the $n-1$-th turn formal dialogue state $d_{t-1} \in \mathcal{D}$; and (4) the \worksheet{} $w \in \mathcal{W}$.

\section{Challenging Conversations}
\label{sec:challenging}
We show challenging conversation snippets where traditional dialog-tree based assistants or LLM-based agents fail in Table~\ref{tab:challenging}.
\begin{table*}[h]
\small
    \centering
    \begin{tabular}{p{3cm}|p{10cm}}
    \toprule
        \textbf{Example} & \textbf{Conversation} \\
        \midrule
        The user initiatives are given priority &  \textbf{Agent: } How much money would you like to transfer to account A?\newline
        \textbf{User:} How much money do I have in my account?\\ \midrule
         Achieving a goal can require completing multiple tasks & \textbf{User: } I want to enroll in CS 101 for 3 units.\newline
         \textbf{Agent: } Thank you for the information. I need your name and student ID.\\ \midrule
         Some fields are needed only for some  users & \textbf{Agent: } Are you an international student? \newline
         \textbf{User: }Yes, I am currently on a study visa, I hope to get a residency card soon.\newline
         \textbf{Agent: }I hope you get your residency soon. Can I get your visa number, please? \\ \midrule
         A single utterance can generate queries to multiple knowledge bases & \textbf{User: } Can you suggest some highly rated NLP courses? Also, when does the next semester start?\newline
         \textbf{Agent: } The have Modeling linguistic complexities it has a rating of 4.6. According to the latest circular, the next semester starts on $18^{\text{th}}$ August. \\ \midrule
         Conversations can be long, and it is difficult to keep track of essential information & \textbf{User: } ``I want to take some AI course for a letter grade and 4 units." \newline
\textit{The user interacts with the agent exploring details for 15 different courses} \newline
\textbf{User}: ``The Math Theory for ML course looks interesting, I will enroll in it." \newline
\textbf{Agent:} That sounds great! I have enrolled you in Math Theory for ML for a letter grade and 4 credit units.\\ \midrule
User queries often contain task and knowledge requests & \textbf{User: }I want to book a romantic restaurant in NYC for valentines day! \newline
         \textbf{Agent: }That sounds lovely! I have found La Laterna, which is a popular restaurant among couples. They are also taking reservations for Valentine's day.\\ \midrule
         Users change their answers frequently & \textbf{User: }I want to fly from New York to Los Angeles on 8th this month.\newline
         \textbf{Agent: } I have found 3 flights for you. ... Which one do you want to fly?\newline
         \textbf{User: } Oh wait, I have a meeting on 8th, I think I will leave on 9th.\\ 
\bottomrule
    \end{tabular}
    \caption{Challenging conversation snippets where traditional dialog-tree based assistants or LLM-based agents fail }
    \label{tab:challenging}
\end{table*}

\section{Difference between STARv2 and the real world domains.}
\label{app:diff_starv2}
We choose three diverse applications with varying complexities. Compared to the 3 StarV2 domains,
these real-world applications are much more complicated, as shown in~\autoref{table:stats}. 
When StarV2 domain is converted into Worksheet specification, it has maximum of 3 Worksheets, they do not have any knowledge base associated with them and contain up to 10 fields.
On the other hand, we test perform user study on domains with up to 7 worksheets, 4 knowledge base declarations, 52 fields and 18 predicates.

\begin{table*}[h]
\centering
\small
\begin{tabular}{@{}lrrrrr@{}}
\toprule
\textbf{Applications}  & \textbf{Task WSs} & \textbf{KB WSs} & \textbf{Fields}  & \textbf{Predicates} & \textbf{Actions} \\ \midrule

StarV2 (Bank)   & 3 & 0 & 10 & 4 & 4 \\
StarV2 (Trip)   & 2 & 0 & 6 & 0 & 2 \\
StarV2 (Trivia) & 2 & 0 & 6 & 0 & 3 \\ \midrule
Restaurant Reservation & 2         & 2              & 19                                                 & 2                      & 3                   \\
Course Assistant       & 4           & 4              & 52                                               & 3                      & 1                   \\
Ticket Submission      & 7            & 1             & 29                                             & 18                     & 2                   \\ 
\bottomrule
\end{tabular}
\caption{Statistics for 3 real-world applications and 3 StarV2 domains with the total number of Task and KB Worksheets, fields, predicates, and actions defined in the respective \worksheet{}s.}
\label{table:stats}
\end{table*}

\section{Baseline Settings}
\label{app:baseline_settings}

\autoref{table:instructions_llm} shows the sample instruction provided to the LLM based on the decision diagram in STARv2 for the banking domain. We set the temperature to 0 and prompt the LLM to choose one of the predefined agent acts provided.

\begin{table*}[h!]
\begin{lstlisting}[basicstyle=\ttfamily\tiny]
1. **Initial Request for Information**:
   - The process begins with a request for the user's **Full Name**, **Account Number**, and **PIN**.
   - If the user is unable to provide these details, they are prompted for **additional information**:
     - **Full Name**
     - **Date of Birth**
     - **Security Answer 1** (e.g., Mother's maiden name)
     - **Security Answer 2** (e.g., Name of childhood pet)

2. **User Information Submission**:
   - If the user provides the requested information, the process continues.
   - If the user fails to provide the required information, they are again prompted, and the process either proceeds 
   or halts based on the completeness of their response.

3. **Request for Fraud Details**:
   - Upon successful submission of initial authentication details, the system requests specific details for the fraud 
   report.

4. **Querying the Bank Fraud Report**:
   - The system queries the **Bank Fraud Report** database to validate the user's information and report the fraud.

5. **Outcome**:
   - **Success**: If the user's details are authenticated successfully, they are informed that their fraud report has 
   been recorded.
   - **Authentication Failure**: If the system cannot authenticate the user's information, they are informed that their
   identity could not be verified, and the process terminates.
\end{lstlisting}
\caption{The instruction provided to the LLM for Bank Domain. Used by both GPT-4 Turbo, Llama 3.1 70B.}
\label{table:instructions_llm}
\end{table*}

\begin{table*}[h!]
\begin{lstlisting}[basicstyle=\ttfamily\tiny]
<|startofinstruction|>
You will be given a dialog state, and user utterance. Your goal is to generate a the next agent act.

The policy for the task is:
{{ policy}}

List of acts:
{{ acts }}

Write the next agent act on a new line inside <agent_act> </agent_act> tags.
<|endofinstruction|>

<|startofinput|>
Dialog state:
{{ dialog_state }}

Previous agent utterance:
{{ previous_agent_utterance }}

User utterance:
{{ user_utterance }}
<|endofinput|>
\end{lstlisting}
\caption{Prompt used for STARv2 zero-shot baselines. Used by both GPT-4 Turbo, Llama 3.1 70B.}
\label{table:llm_baseline_prompt}
\end{table*}

The agent acts for trip domain:
\begin{itemize}
\item "hello": "say hello to the user"
\item "trip\_ask\_travel\_mode": "request Travel Mode from user"
\item "ask\_departure\_location": "request Departure Location from user"
\item "trip\_ask\_arrival\_location": "request Arrival Location from user"
\item "trip\_ask\_departure\_time": "request Departure Time from user"
\item "query": "query api for trip directions"
\item "trip\_inform\_simple\_step\_ask\_proceed": "inform user a simple direction"
\item "trip\_inform\_detailed\_step": "inform user a detailed direction"
\item "trip\_inform\_last\_step\_and\_done": "inform user final direction"
\item "trip\_instructions\_done": "inform user they should be at destination"
\item "anything\_else": "ask the user if there's anything else they need"
\item "goodbye\_1": "say goodbye to the user"
\item "trip\_inform\_nothing\_found": "inform user you weren't able to find any directions"
\item "out\_of\_scope": "tell user that you're not sure what they'd like to do"

\end{itemize}

\section{Experimental Settings}
\label{appendix:experimental_settings}

\subsection{\system{} Agent's Settings}
\paragraph{STARv2} \system{} uses GPT-4 Turbo for semantic parsing and response generation.
Whereas, we use the function calling implementation in langchain using the \texttt{tool} package.
For Llama 3.1 70B Instruct experiments, we use TogetherAI endpoint.

The agent policy for Banking domains with \system can be specified in 9 lines of code in \worksheet{}, as opposed to 31 lines in AnyTOD.

\paragraph{User Study}
For all the user study, \system{} uses OpenAI's GPT-4 Turbo (version gpt-4-turbo-2024-04-09) for semantic parsing (\parserem{}) and response generation. We utilize the SUQL query language~\citep{liu2023suql} to handle queries on both structured data and unstructured data, which augments SQL with free text constructs. The underlying SUQL system uses GPT 3.5 Turbo (version gpt-3.5-turbo-0125) for Restaurant Reservations and Ticket Submission assistants . We use Azure OpenAI for these model accesses. 

\paragraph{Course Enrollment Assistant Knowledge Parser:}
\label{appendix:knowledge_parser_setting}
To handle multiple tables, we realized that using the original SUQL semantic parsing fails often, which can provide worse experience to the users.
Therefore, we implement an agentic approach for performing semantic parsing.
We provide the agent the following functions:
\begin{itemize}
    \item \texttt{get\_tables\_schema(question)}: Retrieves all the tables and their columns in the SQL database.
    \item \texttt{get\_examples(question)}: Returns relevant examples of SQL queries based on the question.
    \item \texttt{execute\_sql(sql)}: Runs a SQL query and returns a truncated result set for brevity.
    \item \texttt{get\_feedback\_on\_result(sql)}: Gives expert feedback the SQL query result.
    \item \texttt{stop()}: Marks the last executed SQL query as the final answer and ends the process.
\end{itemize}

The semantic parser (\parserem{}) uses a temperature of 0.0, and the response generation module uses a temperature of 0.7.

\subsection{Cost and Latency}
Our system makes two LLM calls per turn: one for semantic parsing and one for response generation. This leads to slightly higher costs and latency, but only 25\% more than the baseline GPT-4 (FC). Additionally, techniques like distillation can be used to lower the cost and inference time as shown by other works.

Specifically, on average, our semantic parsing takes 1.141 seconds, and response generation takes 4.261 seconds. 

\section{Evalution Metrics}
\label{appendix:eval_metric}

\noindent\textbf{Semantic Parsing Accuracy}
For each user turn, we manually inspect the code generated by the semantic parser for correct APIs and Databases and filled fields.
We define the gold Semantic Parsing output ($SP$) as the set of correct API calls ($A$), Database calls ($D$), and fields to fill ($F$).
Then for each user utterance,  $SP = \{s_1, s_2, ..., s_m \}$, where $s_i \in \{A \cup D \cup F\}$ and $m$ is the total number of choices, such that $m= |A \cup D \cup F|$. 
The Semantic Parsing Accuracy (SP Acc) for a system is defined as the number of correct choices in the semantic parser's output divided by $m$.

\noindent\textbf{Execution Accuracy}
We manually inspect each turn to check whether the agent executes the correct API and databases. 
For each agent turn, let the executions be $E =\{e_1, e_2, ..., e_m\}$ where $e_i$ is an API or database call.
We evaluate whether each $e_i$ is a true positive or false positive. We calculate the Execution (Ex Acc)  as the number of true positives divided by the number of true and false positives for all the execution calls in a conversation.

\noindent\textbf{Agent dialog Act Accuracy}
For each turn, we manually inspect whether the agent's dialog act follows the policies provided by the developer
. Formally, for each turn, we define a list of gold acts $a_1, a_2, ..., a_m$ where $a_i \in A$, all the possible agent dialog acts.
The Agent dialog Act Accuracy (DA Acc) for a system is defined as the number of correctly predicted actions divided by $m$. 
For GPT-4 (FC), we map its responses to the equivalent elements in the power set of agent acts.

\noindent\textbf{Goal Completion Rate}
We define Goal Completion Rate (Goal CR) as the user's ability to successfully complete the task with the system's assistance.
with the appropriate parameter values, then the goal is considered completed.
Goal Completion Rate is a binary metric for each conversation, where 1 denotes that the goal was completed, and 0 indicates that the goal was not achieved.

\section{Details on Nemo Guardrails Experiments}
\label{sec:details-on-nemo}

\begin{table}[h!]
\centering
\small
\begin{tabular}{@{}lrrrr@{}}
\toprule
\textbf{Query}  & \textbf{\# 1} & \textbf{\# 2} & \textbf{\# 3} & \textbf{\# 4} \\ \midrule
Program \# 1 & 10/10 & 3/10 & 0/10   & 3/10  \\
Program \# 2 & 10/10 & 7/10 & 0/10   & 0/10  \\\bottomrule
\end{tabular}
\caption{Success rate of Nemo Guardrails Program \# 1 (Table \ref{rail:1}) and \# 2 (Table \ref{rail:2}) on the basic restaurant booking workflow. Each query is re-run 10 times to account for the probabilistic-nature of LLM-based programs. Experimented are conducted with \texttt{gpt-4-turbo-2024-04-09} as the LLM backbone}
\label{table:nemo-results}
\end{table}

We experimented with 2 Nemo Guardrail ``rail'' programs conceivable to fulfill a basic restaurant booking task-oriented dialog workflow, where the agent needs to ask users to ``slot-fill'' 4 required variables (restaurant name, date of reservation, time of reservation, and number of people in the reservation) before confirming the reservation with user. The two programs to the best of our knowledge to complete this task are shown in Table \ref{rail:1} and \ref{rail:2}. We experimented with some simple, single-turn user queries (imitating how a user would start a conversation):

\begin{enumerate}
    \item ``Hey I'd like to book a restaurant''
    \item ``Hey I'd like to book Sanju's Bistro \& Grill''
    \item ``Hey I'd like to book Sanju's Bistro \& Grill at 5 PM on 10/1''
    \item ``Hey I'd like to book a restaurant at 5 PM on 10/1''
\end{enumerate}

Table \ref{table:nemo-results} shows the success rate of these 2 programs on these 4 queries across 10 independent runs, where we define a successful response as one that continues the conversation and asks for at least one extra variable. While both program performs well on the simplest Query \# 1, they all fail for the rest, where the LLM refuses to slot-fill additional variables and instead respond with outputs such as ``"I'm here to provide information and assist with general inquiries, but I don't have the capability to make real-time bookings. I recommend checking with the restaurant directly or using a booking service to secure your reservation.''

\begin{table*}[h!]
\begin{lstlisting}[basicstyle=\ttfamily\tiny]
define user express greeting
  "hello"
  "hi"

define bot express greeting
  "Hello there!"
  "Hi!"

define user request book restaurant
  "I want to book a restaurant"

define bot request which restaurant
  "Which restaurant would you like to book?"

define bot request what time
  "What time would you like to book?"

define bot request what date
  "Which date would you like to book?"

define bot request how many ppl
  "How many people are you booking for?"

define bot express confirm
  "Ok. Confirming you are booking $restaurant on $date at $time for $num_ppl, is that correct?"

define flow book_restaurant
  user request book restaurant

  bot request which restaurant

  user provide which restaurant

  # Extract the desired restaurant from the user's request as a string literal. If not specified, set as `None`.
  $restaurant = ...

  bot request what date

  user provide what date

  # Extract the desired date from the user's request as a string literal. If not specified, set as `None`.
  $date = ...

  bot request what time

  user provide what time

  # Extract the desired time of the day from the user's request as a string literal. If not specified, set as `None`.
  $time = ...

  bot request how many ppl

  user provide how many ppl

  # Extract the desired number of people from the user's request as an integer. If not specified, set as `None`.
  $num_ppl = ...

  bot express confirm
\end{lstlisting}
\caption{Nemo Guardrail program \#1. The main ``flow'' is defined under \texttt{book\_restaurant}. The reservation workflow is implemented as pairs of ``bot request'', ``user provide'', and variable extraction (e.g. \texttt{\$restaurant = ...}). The same logic is repeated 4 times for the 4 relevant variables. At the end, the bot expresses confirmation of restaurant booking and relays relevant variables to the user.}
\label{rail:1}
\end{table*}

\newpage

\begin{table*}[h!]
\begin{lstlisting}[basicstyle=\ttfamily\tiny]
define user express greeting
  "hello"
  "hi"

define bot express greeting
  "Hello there!"
  "Hi!"

define bot request which restaurant
  "Which restaurant would you like to book?"

define bot request what time
  "What time would you like to book?"

define bot request what date
  "Which date would you like to book?"

define bot request how many ppl
  "How many people are you booking for?"

define bot express confirm
  "Ok. Confirming you are booking $restaurant on $date at $time for $num_ppl, is that correct?"

define flow book_restaurant
  # Extract the desired restaurant from the user's request as a string literal. If not specified, set as `None`.
  $restaurant = ...

  # Extract the desired date from the user's request as a string literal. If not specified, set as `None`.
  $date = ...

  # Extract the desired time of the day from the user's request as a string literal. If not specified, set as `None`.
  $time = ...

  # Extract the desired number of people from the user's request as an integer. If not specified, set as `None`.
  $num_ppl = ...

  while not $restaurant:
    bot request which restaurant

    user provide which restaurant

    # Extract the desired restaurant from the user's request as a string literal. If not specified, set as `None`.
    $restaurant = ...

  while not $date:
    bot request what date

    user provide what date

    # Extract the desired date from the user's request as a string literal. If not specified, set as `None`.
    $date = ...

  while not $time:
    bot request what time

    user provide what time

    # Extract the desired time of the day from the user's request as a string literal. If not specified, set as `None`.
    $time = ...

  while not $num_ppl:
    bot request how many ppl

    user provide how many ppl

    # Extract the desired number of people from the user's request as an integer. If not specified, set as `None`.
    $num_ppl = ...

  bot express confirm
\end{lstlisting}
\caption{Nemo Guardrail program \#2. The main ``flow'' is defined under \texttt{book\_restaurant}. The 4 relevant variables are declared up-front, and 4 while loops are used to guide the agent to prompt for any unfilled variables. At the end, the bot expresses confirmation of restaurant booking and relays relevant variables to the user.}
\label{rail:2}
\end{table*}

\section{Prompt for \parserem{} Contextual Semantic Parser and Response Generator}
\label{appendix:prompt-for-csp-and-rg}
The prompt for CSP is present in Table~\ref{prompt:csp} and the example prompt is present in Table~\ref{example:csp}. The prompt for RG is present in Table~\ref{prompt:rg} and the example prompt is present in Table~\ref{example:rg}

\begin{table*}[h!]
\begin{lstlisting}[basicstyle=\ttfamily\tiny]
# System Prompt
You are a semantic parser. Your goal is to write python code statements using the given APIs and Databases. 
Plan your response first, then write the code.

Today's date is {{ date }} and the day is {{ day }}.

These are the APIs available to you:
{{ APIs }}

You will be given:
- The state of the conversation.
- The agent's action.
- The last turn of the conversation.

Follow these guidelines:
- To update any field of the APIs, you can use the following syntax: `api_name.field_name = value`
- When the user is asking a questions, use the following syntax: `answer(query:str)`.
- Fill the fields of the APIs with the information provided by the user. Do not asssume any value, you can leave it empty.
- Update the state if the user is updating information. Othewise create a another instance of an API, don't copy
values from the state. Always use the new information.
- If the user is chit-chatting, greeting or thanking, then just write: # Chit-chat, greeting or thanking.
- Do not create lists of multiple answer instances. Write them in separate lines if the user is asking multiple questions.
- Never assign `Main` to a variable directly. Always update the `main` instance in the state.

Here are some examples:
{{ examples }}

# User Prompt
State:
```
{{ state }}
```
Agent Action:
```
{{ agent_actions }}
```

Previous conversation turns:
Agent: {{ agent_utterance }}
User: {{ user_utterance }}

User Target:
\end{lstlisting}
\caption{Prompt template used for CSP}
\label{prompt:csp}
\end{table*}

\begin{table*}[h!]
\begin{lstlisting}[basicstyle=\ttfamily\tiny]
Example:  To update any field of the APIs, you can use the following syntax: `api_name.field_name = value`
State:
```
course = Course(course_name = 'CS 224C', grade_type = 'Letter')
courses_to_take = CoursesToTake(course_0_details = course)
main = Main(courses_to_take = courses_to_take)
```
Agent Action:
```
[
  AskField(course, course_num_units, Number of credit units for the course)
]
```

Previous conversation turns:
Agent: How many units would you like to take for CS 224C?
User: I would take 3 units

User Target:
```
course.course_num_units=3
```
\end{lstlisting}
\caption{Sample \texttt{example} provided to CSP}
\label{example:csp}
\end{table*}

\newpage

\begin{table*}[h!]
\begin{lstlisting}[basicstyle=\ttfamily\tiny]
# System Prompt
{{ description }}

You are talking to a student about course enrollments. You will be given a list of agent actions and you have to
use them to respond to the user.

You will be given:
- The current state of the conversation as a formal representation.
- A list of actions that you have to perform.
- Conversation between the user and the agent.

Today's date is {{ date }} and the day is {{ day }}.

These are the actions that you can perform:
- `AskField(worksheet, field, field_description)`: Ask the user for the value of the field in the worksheet
with the description of the field.
- `AskForConfirmation(worksheet)`: Ask the user if they want to perform action of the worksheet using the values 
of the fields in the given worksheet. 
- `Report(query, answer)`: Report the answer of the query to the user. The answer could be an object 
of a class or a dictionary.
- `ProposeWorksheet(worksheet, parameters)`: Propose the worksheet to the user with parameters as the 
values of the fields in the worksheet.
- `AskForFieldConfirmation(worksheet, field, value)`: Ask the user if they want to confirm the
value of the field in the worksheet.

You will be given:
- state of the conversation as a formal representation.
- a list of actions that you have to perform.
- conversation between the user and the agent.

Follow these guidelines:
- Always perform the actions that are in the Agent Action list. The response should be the same
as the description of the agent action.
- DO NOT leave any action unperformed.
- Never respond with "Give me a moment" or "I am thinking". Always respond with the information you have.

Here are some examples:
{{ examples }}

# User Prompt
State:
```
{{state}}
```
Agent Action:
{{agent_acts}}

Agent: {{ prior_agent_utterance }}
User: {{ user_utterance }}
Agent:
\end{lstlisting}
\caption{Prompt template used for RG}
\label{prompt:rg}
\end{table*}

\newpage

\begin{table*}[h!]
\begin{lstlisting}[basicstyle=\ttfamily\tiny]
Example: If the answer.result is empty, then first tell the user what you searched and tell the user 
that you don't have any information available.
State:
```
answer = answer('What are the other courses available?', 
sql="SELECT title, course_codes, description FROM courses WHERE answer(description,
'is this a programming course?') = 'Yes' OFFSET 3;")
answer.result = []
main = Main()
```

Agent Action:
```
[
  "Report(answer, answer.result)",
  "AskField(main, courses_to_take, The courses that the student wants to take)"
]
```

Agent: Unfortunately, there are no reviews available for the course "Domain-Specific Programming Models 
and Compilers (CS 343D)" at this time. Can I assist you with anything else regarding this course or any other 
courses you're interested in?
User: What about the other courses?
Agent: I searched for other programming courses but I couldn't find any information. Could you please 
provide me with the details of the courses you want to take?
\end{lstlisting}
\caption{Sample \texttt{example} provided to RG}
\label{example:rg}
\end{table*}

\newpage

\section{Sample Conversations}
\label{app:examples}
\subsection{Example Conversations}

Here are examples of good and bad conversations for the \system{} system and the GPT system across all three applications. University names, services, and some classes have been obfuscated.

\subsubsection{\system{} Course Enroll Good Example}
\textbf{User: }Find me programming courses that will teach me object oriented programming\\
\textbf{Bot: }I searched for programming courses that specifically teach object-oriented programming, but unfortunately, I couldn't find any available courses. Could you please provide me with the details of the courses you want to take?\\
\textbf{User: }Provide me details about CS108\\
\textbf{Bot: }I searched for details about CS108. Here is the information:
- **Course Title**: Object-Oriented Systems Design
- **Course Codes**: CS 108
- **Description**: Software design and construction in the context of large OOP libraries. Taught in Java. Topics include OOP design, design patterns, testing, graphical user interface (GUI) OOP libraries, software engineering strategies, and approaches to programming in teams. Prerequisite: CS 107.
- **Average Hours Spent**: 16.29 hours per week.

Could you please provide me with the details of the courses you want to take?\\
\textbf{User: }How does this class compare to CS124?\\
\textbf{Bot: }I compared CS108 and CS124 based on their average ratings. Here's what I found:
- **CS108 - Object-Oriented Systems Design** has an average rating of 4.13.
- **CS124 - From Languages to Information** has an average rating of 4.58.

It appears that CS124 has a higher average rating compared to CS108. 

Could you please provide me with the details of the courses you want to take?\\
\textbf{User: }Am I well prepared for "CS Course" after taking CS124?\\
\textbf{Bot: }I searched to see if taking CS 124 prepares you for CS 224N, but I couldn't find any direct information indicating that CS 124 is a prerequisite for CS 224N. Could you please provide me with the details of the courses you want to take?\\
\textbf{User: }I would like to take classes related to the movie industry and movie making, such as graphics and computer vision. Can you give me some recommendations?\\
\textbf{Bot: }I searched for courses related to the movie industry and movie making, focusing on graphics and computer, and found a couple of interesting options:
1. **Topics in Computer Graphics: Agile Hardware Design (CS 448H)** - This course covers various topics which change each quarter, including computational photography, data visualization, character animation, virtual worlds, graphics architectures, and advanced rendering. More details can be found at [University's graphics course offerings](http://graphics.university.edu/courses). This course can be repeated for credit.
2. **Topics in Computer Graphics (CS 448)** - Similar to CS 448H, this course also explores topics like computational photography, data visualization, character animation, and more, with details available on the [course website](http://graphics.university.edu/courses). This course is also repeatable for credit.

Could you please provide me with the details of the courses you want to take?\\
\textbf{User: }How many units is CS 448? And when is it offered?\\
\textbf{Bot: }I looked up the course CS 448, specifically 'Data Visualization', and found that it is offered as a 4-unit course. The class sessions are scheduled on Mondays and Wednesdays from 11:30 AM to 12:50 PM. Could you please provide me with the details of the courses you want to take?\\
\textbf{User: }What quarter is it offered?\\
\textbf{Bot: }I checked the quarter in which CS 448B 'Data Visualization' is offered. It is offered in the autumn quarter. Could you please provide me with the details of the courses you want to take?\\
\textbf{User: }Let's enroll in CS448\\
\textbf{Bot: }Could you please specify the desired grading basis for CS 448B 'Data Visualization'? The options are: Credit/No Credit, Letter.\\
\textbf{User: }Letter\\
\textbf{Bot: }Thank you for confirming the grading basis. Could you please specify the number of units you want to take for the course CS 448B 'Data Visualization'?\\
\textbf{User: }3 units\\
\textbf{Bot: }Thank you! Now, could you please provide the details for the second course you would like to take? This includes the course name, grade type, and number of units.\\
\textbf{User: }Let's also take CS147 for 5 units in the fall\\
\textbf{Bot: }Thank you for the information on the second course, CS147. Do you wish to take more courses?\\
\textbf{User: }No, that is all. Can you please tell me what I am currently enrolled in?\\
\textbf{Bot: }I have the details for the courses you've selected so far. Could you please confirm all the course details before we proceed?

- **First Course:**
  - Name: CS 448B
  - Grade Type: Letter
  - Number of Units: 3
- **Second Course:**
  - Name: CS147
  - Grade Type: Letter
  - Number of Units: 5

Can you confirm all the course details before saving?\\
\textbf{User: }Yes that's correct!\\
\textbf{Bot: }Thank you for confirming the course details. Could you please provide me with some additional details about the student, such as the name, student ID, and student email address?\\
\textbf{User: }My name is Roger Corman, ID rogerc, and email roger@university.edu\\
\textbf{Bot: }Thank you for providing your details, Roger. Could you please confirm if all the information you want to submit is correct?\\
\textbf{User: }Yes that is correct\\
\textbf{Bot: }Thank you for confirming. Your course enrollment has been successfully processed. Your transaction ID is 4b087961-b779-4958-a205-9a0938e4cbd0. If you have any other questions or need further assistance, feel free to ask!\\

\subsubsection{\system{} Course Enroll Bad Example}
\textbf{User: }Compile a list of computer vision courses available \\
\textbf{Bot: }I searched for computer vision courses available and found the following:

1. **Computer Vision: From 3D Reconstruction to Recognition** (CS 231A): An introduction to the concepts and applications in computer vision. Topics include cameras and projection models, low-level image processing methods such as filtering and edge detection; mid-level vision topics such as segmentation and clustering; shape reconstruction from stereo, as well as high-level vision tasks such as object recognition, scene recognition, face detection, and human motion categorization.

2. **Deep Learning for Computer Vision** (CS 231N): This course is a deep dive into details of the neural-network-based deep learning methods for computer vision. It covers learning algorithms, neural network architectures, and practical engineering tricks for training and fine-tuning networks for visual recognition tasks.

3. **Computer Vision and Image Analysis of Art** (CS 231C): Presents the application of rigorous image processing, computer vision, machine learning, computer graphics, and artificial intelligence techniques to problems in the history and interpretation of fine art paintings, drawings, murals, and other two-dimensional works, including abstract art.

4. **Generative Adversarial Networks** (CS 236G): Generative Adversarial Networks (GANs) have rapidly emerged as the state-of-the-art technique in realistic image generation. This course presents theoretical intuition and practical knowledge on GANs, from their simplest to their state-of-the-art forms.

5. **AI for Social Good** (CS 21SI): Students will learn about and apply cutting-edge artificial intelligence techniques to real-world social good spaces. The class will focus on techniques from machine learning and deep learning, including regression, neural networks, convolutional neural networks (CNNs), and recurrent neural networks (RNNs).

Could you please provide me with the details of the courses you want to take?\\
\textbf{User: }What is available in summer quarter?\\
\textbf{Bot: }I searched for computer vision courses available in the summer quarter, but unfortunately, I couldn't find any information on computer vision courses being offered during that time. Could you please provide me with the details of any other courses you are interested in taking this summer?\\
\textbf{User: }which ones do not involve UNet?\\
\textbf{Bot: }I searched for summer quarter courses that do not involve UNet, but I couldn't find any information on such courses. Could you please provide me with the details of any other courses you are interested in taking this summer?\\
\textbf{User: }dont restrict yourself to  summer\\
\textbf{Bot: }Understood, without restricting to a specific quarter, let's look into the courses again. Could you please provide me with the details of the courses you are interested in taking?\\
\textbf{User: }i am interested in computer vision courses which do not involve UNet\\
\textbf{Bot: }I searched for computer vision courses that do not involve UNet, but unfortunately, I couldn't find any information. Could you provide me with more details or any specific requirements for the computer vision courses you are interested in taking?\\
\textbf{User: }which ones of the above involve less than 15 hours effort per week\\
\textbf{Bot: }I found several computer vision courses that involve less than 15 hours of effort per week. Here they are:

1. **Computer Vision and Image Analysis of Art (CS 231C):** This course focuses on applying image processing, computer vision, machine learning, and other techniques to problems in the history and interpretation of fine art. It covers the analysis of brushstrokes, medium, artists’ working methods, and more. Recommended prerequisites include CS 131 or EE 168.

2. **Generative Adversarial Networks (CS 236G):** This course offers theoretical and practical knowledge on Generative Adversarial Networks (GANs), covering their applications in realistic image editing, medicine, and visualizing scenarios of climate change. Prerequisites include linear algebra, statistics, CS106B, plus a graduate-level AI course such as CS230, CS229, or CS221.

3. **Computer Vision: Foundations and Applications (CS 131):** This course introduces fundamental concepts in image processing and various applications of Computer Vision in automotive, healthcare, and other sectors. It requires familiarity with Python, Calculus, \& Linear Algebra.

4. **AI-Assisted Care (CS 337, MED 277):** Focused on applying computer vision and machine learning technologies in healthcare, this course offers research projects on care for seniors, surgical quality analysis, and more. It is open to students in both the school of medicine and the school of engineering.

5. **Computer Vision for Education and Social Science Research (CS 432, EDUC 463):** This course aims to provide education and social science researchers with the knowledge to apply computer vision algorithms in their work. It requires familiarity with Python and some experience with data.

Could you please provide me with the details of the courses you are interested in taking?\\
\textbf{User: }Which of the above has more grad students than undergrads?\\
\textbf{Bot: }I searched for courses with a minimum of 100 units, as a possible indicator of graduate-level courses, but couldn't find any information on which of the courses has more grad students than undergrads. Could you please provide me with the details of the courses you are interested in taking?\\
\textbf{User: }which of the computer vision courses you mentioned has no final exam and one project?\\
\textbf{Bot: }I searched for computer vision courses that have no final exam, one project, are graded as 'Letter or Credit/No Credit', and have exactly one prerequisite course, but I couldn't find any information. Could you please provide me with the details of the courses you want to take?\\
\textbf{User: }I will take CS131. Summarize the course content\\
\textbf{Bot: }The course CS131 focuses on Computer Vision technologies, which are revolutionizing sectors such as automotive, healthcare, manufacturing, and agriculture. It delves into how technologies enable robots to navigate and perform tasks, search engines to index vast amounts of images and videos, algorithms to diagnose diseases from medical images, and smart cars to drive safely. The course covers fundamental principles and applications of Computer Vision, introducing concepts in image processing and leading students through projects to implement advanced algorithms. Optional discussion sections are available on Fridays. Prerequisites include familiarity with Python, Calculus, and Linear Algebra. Could you please provide me with the details of any other courses you would like to take?\\
\textbf{User: }Yes, I am also thinking about taking CS432. How is the content different from CS131?\\
\textbf{Bot: }I'm sorry, but there seems to be a misunderstanding. The course CS432 was not listed in the results I provided earlier. However, I can share details about other courses that were mentioned:

1. **Hap.py Code: The Python Programming Language (CS 41)** - This course covers the fundamentals and contemporary usage of the Python programming language, focusing on developing best practices in writing Python and exploring the extensible and unique parts of the Python language. Topics include Pythonic conventions, data structures, powerful built-ins, and Python libraries. Towards the end, students will work on developing a significant Python project.

2. **CS + Social Good Studio: Implementing Social Good Projects (CS 52)** - This course is a continuation of CS51, focusing on applying scalable technical frameworks, methods to measure social impact, tools for deployment, user acquisition techniques, and growth/exit strategies to build a sustainable infrastructure around a product idea.

3. **Dissecting The Modern Computer (CS 80E)** - Offers a high-level, accessible introduction to computer architecture through the RISC-V ISA, covering topics from simple digital circuits, assembly, processors, memory systems, and bonus topics like GPUs.

4. **How to Make VR: Introduction to Virtual Reality Design and Development (CS 11SI)** - A hands-on course where students will design and develop virtual reality applications using the Unity game engine, focusing on design best practices and the creation pipeline for VR applications.

5. **Topics in Computer Networks (CS 344)** - Often referred to as "Build an Internet Router," this course provides hands-on experience in building the hardware and software of a high-performance network system, including designing a control plane in C and the data plane in the P4 language.

Could you please provide me with the details of any other courses you would like to take?\\
\textbf{User: }Show me courses for spring quarter which require calculus knowledge\\
\textbf{Bot: }I found some courses for the spring quarter that require calculus knowledge:
1. **Introduction to Probability for Computer Scientists (CS 109)** - Offered on Monday, Wednesday, and Friday from 3:00 PM to 4:20 PM. Instructors include Chris Piech and Jerry Cain.
2. **Natural Language Processing with Deep Learning (CS 224N, LINGUIST 284, SYMSYS 195N)** - Offered on Tuesday and Thursday from 4:30 PM to 5:50 PM. Instructors are Diyi Yang and Tatsunori Hashimoto. Another offering by Christopher Manning does not have specified days or times. 

Could you please provide me with the details of the courses you would like to take?\\
\textbf{User: }I'm taking "CS Course"\\
\textbf{Bot: }Great choice! For "CS Course", could you please specify the desired grading basis? Options are: Credit/No Credit, Letter.\\
\textbf{User: }Letter\\
\textbf{Bot: }Could you also tell me how many units you want to take for this course?\\
\textbf{User: }what is the maximum credits for this?\\
\textbf{Bot: }The maximum number of units you can take for the course CS 224N is 4. How many units would you like to take for this course?\\
\textbf{User: }3\\
\textbf{Bot: }Could you please provide the details for the second course you would like to take, such as the name, grade type, and units?\\
\textbf{User: }I'm taking CS131\\
\textbf{Bot: }What is the desired grading basis for CS131? The options are: Credit/No Credit, Letter.\\
\textbf{User: }Same as the previous one\\
\textbf{Bot: }Could you please specify the number of units you want to take for CS131?\\
\textbf{User: }same as that for "CS Course"\\
\textbf{Bot: }Could you please confirm if you would like to take more courses?\\
\textbf{User: }no\\
\textbf{Bot: }Great, let's review and confirm the details of the courses you've selected:
- **First Course:**
  - Name: CS 224N
  - Grade Type: Letter
  - Number of Units: 3
- **Second Course:**
  - Name: CS 131
  - Grade Type: Letter
  - Number of Units: 3

Can you confirm all the course details before we proceed?\\
\textbf{User: }change the grade type for "CS Course" to credit/no credit\\
\textbf{Bot: }Certainly! I will update the grade type for CS 224N to Credit/No Credit. Here are the updated course details:
- **First Course:**
  - Name: CS 224N
  - Grade Type: Credit/No Credit
  - Number of Units: 3
- **Second Course:**
  - Name: CS 131
  - Grade Type: Letter
  - Number of Units: 3

Can you confirm all the course details before we proceed?\\
\textbf{User: }yes\\
\textbf{Bot: }Great! Now, could you please provide me with the details about yourself such as your name, student ID, and student email address?\\
\textbf{User: }Tony, 6894536, tony@university.com\\
\textbf{Bot: }Thank you, Tony. Could you also tell me if you are an international student?\\
\textbf{User: }no\\
\textbf{Bot: }Great, thank you for confirming that. Before we proceed, could you please confirm all the fields you want to submit for your course enrollment?\\
\textbf{User: }what are the fields?\\
\textbf{Bot: }The fields you need to confirm for your course enrollment are as follows:
- **First Course:**
  - Name: CS 224N
  - Grade Type: Credit/No Credit
  - Number of Units: 3
- **Second Course:**
  - Name: CS131
  - Grade Type: Letter
  - Number of Units: 3
- **Student Information:**
  - Name: Tony
  - Student ID: 6894536
  - Email Address: tony@university.com
  - International Student: No

Could you please confirm if all these details are correct?\\
\textbf{User: }yes\\
\textbf{Bot: }Thank you for confirming the details. Your course enrollment has been successfully processed. Your transaction ID is 6ccfa962-ed48-416e-a3bd-017281733e4c. If you have any more questions or need further assistance, feel free to ask!\\

\subsubsection{\system{} Ticket Submission Good Example}
\textbf{User: }II cannot find my AP credits in my transcript\\
\textbf{Bot: }Could you please specify when the test scores were submitted?\\
\textbf{User: }May 2024\\
\textbf{Bot: }Thank you for providing the submission date. Could you please provide any additional details or specifics about the issue that you would like to add?\\
\textbf{User: }They were for AP psychology\\
\textbf{Bot: }Thank you for providing the details. Could I also get your full name to proceed with addressing your issue regarding the AP psychology credits?\\
\textbf{User: }Jane Doe\\
\textbf{Bot: }Great! Before I submit the ticket, can you confirm the details?
- Student Task: External Test Credits
- Specific Issue: Missing or Incorrect Record
- Test Issues: Credit not posted
- Test Type: Advanced Placement (AP) Scores
- Time of Test Score Submission: May 2024
- Extra Details: They were for AP psychology
- Full Name: Jane Doe\\
\textbf{User: }Correct\\
\textbf{Bot: }Your request has been successfully submitted. Here are the details:
- **Student Task:** External Test Credits
- **Specific Issue:** Missing or Incorrect Record
- **Test Issues:** Credit not posted
- **Test Type:** Advanced Placement (AP) Scores
- **Time of Test Score Submission:** May 2024
- **Extra Details:** They were for AP psychology
- **Full Name:** Jane Doe

Your transaction ID is 38be85b0-da27-4a45-926c-c300389a1b3d. Please keep this for your records. If you have any further questions or need additional assistance, feel free to ask!\\

\subsubsection{\system{} Ticket Submission Bad Example}
\textbf{User: }Hi! I am not able to view my AP credits in my transcript. Are you able to assist with this?\\
\textbf{Bot: }Certainly! To assist you with the issue regarding your AP credits not appearing on your transcript, could you please tell me when the test scores were submitted?\\
\textbf{User: }Last week\\
\textbf{Bot: }Thank you for providing the submission time. Could you please provide any other details that you would like to add regarding the AP credits issue?\\
\textbf{User: }I need them to appear so I don't have to take my gen-eds\\
\textbf{Bot: }Can I get your full name?\\
\textbf{User: }Dave Malloy\\
\textbf{Bot: }Great! Before I submit the ticket, can you confirm the details?

- Student Task: External Test Credits
- Specific Issue: Missing or Incorrect Record
- Test Issues: Credit not posted
- Test Type: Advanced Placement (AP) Scores
- Time of Test Score Submission: Last week
- Extra Details: I need them to appear so I don't have to take my gen-eds
- Full Name: Dave Malloy\\
\textbf{User: }Yeah, and can you tell them it's urgent? I need to know what classes to register for ASAP\\
\textbf{Bot: }Yes, I will make sure to emphasize the urgency of your situation. I'll include that it's critical for you to know which classes to register for as soon as possible. Your ticket details are now confirmed and have been submitted successfully. Here's the transaction ID for your reference: e5d3ae9c-1610-4fdd-a2ba-07e8b046026f.

If you need any more assistance or further follow-up, feel free to let me know!\\
\textbf{User: }I also need to submit for a leave of absence. Where can I do that?\\
\textbf{Bot: }Could you please specify the issues you are facing with your leave of absence? Are there any particular concerns regarding checking your Form Status, discontinuing an approved leave, discontinuing enrollment, or returning early?\\
\textbf{User: }I just need to know how to do it\\
\textbf{Bot: }Thank you, Dave Malloy. Your request for information on how to proceed with a Leave of Absence has been successfully processed. Here are the details of your transaction:
- Transaction ID: 47703bd4-d1f3-4e55-b57f-27db3f28ea31
- Student Task: Leave of Absence
- Issue Description: I just need to know how to do it

Please let me know if there is anything else you need assistance with.\\
\textbf{User: }You didn't answer my question\\
\textbf{Bot: }To enroll in classes, you need to follow these steps:

1. Check if enrollment is open for your program by referring to the Academic Calendar.
2. Ensure there are no enrollment holds on your account. Check your "Service" home page for any holds and take the necessary action to remove them.
3. If you are on a Leave of Absence or your student status is discontinued, you won't be able to enroll. Refer to the pertinent resources provided for more information.

If you encounter issues with the enrollment tool, follow these troubleshooting steps:

1. Remove any remaining "planned" classes.
2. Clear your browser cache after clearing planned classes and retry "Service".
3. Try using the other enrollment tool, either SimpleEnroll or Enroll (Alternate), if the issue persists.

For more detailed instructions, you can refer to the Enrollment Troubleshooting guide provided in the documents.

Please let me know if you need further assistance or have any other questions!\\
\textbf{User: }I don't need to enroll, I need to get a leave of absence\\
\textbf{Bot: }Could you provide more details relating to your request for the status of the Leave of Absence form? This will help me assist you better.\\
\textbf{User: }My mom is sick\\
\textbf{Bot: }I'm sorry to hear about your mom. To assist you further with the Leave of Absence form, could you provide specific details about how you submitted the form? For example, was it via email, in person, or through a staff member?\\
\textbf{User: }In person\\
\textbf{Bot: }When exactly did you submit the Leave of Absence form in person?\\
\textbf{User: }yesterday\\
\textbf{Bot: }Thank you for confirming that, Dave Malloy. Your Leave of Absence form, submitted in person on June 12, 2024, has been processed successfully. Your transaction ID is 2ca6a01a-2d01-4eb7-943a-93c965a9308f. If you need any further assistance, please let me know!\\
\textbf{User: }Thanks!\\
\textbf{Bot: }You're welcome! If you have any more questions or need further assistance, feel free to ask.\\

\subsubsection{\system{} Restaurant Reservation Good Example}
\textbf{User: }I would like to book a table for dinner in San Francisco \\
\textbf{Bot: }I've found a few options for dinner in San Francisco. Here they are:

1. **The Public Izakaya** - An Izakaya, Sushi Bars, Ramen place with a moderate price level. It's rated 4.5 based on 243 reviews. Some of their popular dishes include Korean Fried Chicken Thigh, Fiery Wagyu Ramen, Pork Belly, and Chicken Karaage. You can reach them at (415) 658-7588. Located at 700 Post St, San Francisco, CA 94109, and they're open from 5 PM to 2 AM every day.

2. **Fang** - A Chinese, Asian Fusion restaurant with a moderate price level. It has a rating of 3.5 based on 1823 reviews. Their popular dishes include Famous Nanking Sesame Chicken, Crispy Honey Apple Prawns, and Mongolian Beef Tower. Contact number is (415) 777-8568. Located at 660 Howard St, San Francisco, CA 94105. Their hours are from 11 AM to 2 PM and 5:30 PM to 8:30 PM throughout the week.

3. **Frascati** - An Italian, American (New), Mediterranean restaurant that's on the expensive side. It has a 4.5 rating from 1273 reviews. Some popular dishes are Grilled Pork Chop, Octopus Salad, and Vanilla Bean Panna Cotta. Their phone number is (415) 928-1406. Located at 1901 Hyde St, San Francisco, CA 94109. They're open from 5 PM to 9 PM (till 9:30 PM on Thursday and Friday).

Could you confirm which restaurant you'd like to book?\\
\textbf{User: }Can you give me options for pizza in San Francisco with a patio \\
\textbf{Bot: }I've found two great pizza restaurants in San Francisco that have a patio:

1. **Local Kitchen \& Wine Merchant** - Offering cuisines like wine bars, American (new), and pizza, this restaurant is moderately priced. It has a rating of 3.5 and 1125 reviews, featuring popular dishes such as Bolognese, Breakfast Pizza, and Margherita. You can contact them at (415) 777-4200. Located at 330 1st St, San Francisco, CA 94105. Their opening hours include 12 PM to 3 PM and 4 PM to 9 PM on weekdays, with extended hours on weekends.

2. **Ragazza** - An Italian pizza place, also moderately priced. It enjoys a rating of 4.0 based on 741 reviews, with favorites like Amatriciana, Moto, and Bianca. Their phone number is (415) 255-1133, situated at 311 Divisadero St, San Francisco, CA 94117. Open from 5 PM to 9 PM daily.

Could you confirm which one you'd like to book?\\
\textbf{User: }2\\
\textbf{Bot: }Great choice! Ragazza it is. Could you please provide the date for your reservation?\\
\textbf{User: }7/5/24\\
\textbf{Bot: }Fantastic, you've selected July 5th, 2024 for your reservation at Ragazza. At what time would you like to book your table?\\
\textbf{User: }2 pm\\
\textbf{Bot: }Great! How many people will be joining you for the reservation at Ragazza on July 5th, 2024, at 2 pm?\\
\textbf{User: }3\\
\textbf{Bot: }Fantastic! Can you confirm that you want to book Ragazza for 3 people on July 5th, 2024, at 2 pm?\\
\textbf{User: }Can I change it to 1 pm\\
\textbf{Bot: }Absolutely! Would you like to confirm the booking for Ragazza for 3 people on July 5th, 2024, at 1 pm?\\
\textbf{User: }Can I change it to 7 people\\
\textbf{Bot: }Of course! Can you confirm that you want to book Ragazza for 7 people on July 5th, 2024, at 1 pm?\\
\textbf{User: }Can I change the date to 7/6/24\\
\textbf{Bot: }Absolutely, I've updated your reservation details. Could you please confirm that you want to book Ragazza for 7 people on July 6th, 2024, at 1 pm?\\
\textbf{User: }Can you tell them it's my birthday \\
\textbf{Bot: }Of course! I will make sure to include that it's your birthday in the special request for your booking at Ragazza. Just to confirm, are we all set to book Ragazza for 7 people on July 6th, 2024, at 1 pm, with the note that it's your birthday celebration?\\
\textbf{User: }Can we sit outside\\
\textbf{Bot: }Absolutely! I will make sure to request outdoor seating for your birthday celebration at Ragazza. Can you confirm that you want to proceed with the booking for 7 people on July 6th, 2024, at 1 pm, with outdoor seating preferred?\\
\textbf{User: }Can you let them know I have a peanut allergy\\
\textbf{Bot: }Certainly! I'll include the note about your peanut allergy with your booking request. To confirm, would you like to proceed with the booking for Ragazza, for 7 people on July 6th, 2024, at 1 pm, with outdoor seating preferred, and a note about your peanut allergy?\\
\textbf{User: }Yes\\
\textbf{Bot: }Your reservation at Ragazza for 7 people on July 6th, 2024, at 1 pm has been successfully booked. Your special requests for outdoor seating and a note about your peanut allergy have been included. Your booking ID is e3a5f9dd-1432-4f1a-9d2b-16886ad79baf. Have a wonderful birthday celebration!\\

\subsubsection{\system{} Restaurant Reservation Bad Example}
\textbf{User: }Hi, I'd like to book a table for dinner.\\
\textbf{Bot: }To help you with your booking, could you please tell me which restaurant you'd like to book a table at?\\
\textbf{User: }Can you suggest some French restaurants nearby SF?\\
\textbf{Bot: }I've found some French restaurants nearby San Francisco for you. Here are a few options:

1. **Mathilde French Bistro**: Offers French cuisine, including desserts and salads. It's an expensive option with a 4.5-star rating based on 345 reviews. Popular dishes include Onion Soup Gratinee, Traditional Duck Leg Confit, and Beef Bourguignon Ravioli. Located at 315 5th St, San Francisco, CA 94107. Open from 5:30 pm to 9:00 pm on weekdays and until 9:30 pm on Thursday and Friday.

2. **Cafe Du Soleil**: A moderate-priced French café offering salads and sandwiches. It has a 4.5-star rating from 437 reviews. Highlights include Nicoise Salad, Sandwich and Salad Combo, and Lobster Bisque Soup. Found at 345 3rd St, San Francisco, CA 94107. Open from 10:00 am to 4:00 pm on weekdays, and until 5:00 pm on Wednesday.

3. **Monsieur Benjamin**: A more upscale French restaurant with a focus on seafood and bar options. It has a 4.0-star rating and 1086 reviews. Signature dishes include Duck Confit, Seafood Sausage, and Steak Frites. The address is 451 Gough St, San Francisco, CA 94102. Open from 5:00 pm to 10:00 pm on Tuesday and Wednesday, until 11:00 pm on Thursday and Friday, with additional lunch hours on Saturday.

Which one are you interested in booking?\\
\textbf{User: }Which one is most Lavish \\
\textbf{Bot: }Based on the price and the offerings, Mathilde French Bistro and Monsieur Benjamin are both considered expensive, which generally indicates a more lavish dining experience. Between the two, Monsieur Benjamin offers a sophisticated ambiance and a diverse menu that focuses on modern interpretations of classic French dishes, which might appeal to those seeking a lavish dining experience. Would you like to proceed with booking a table at Monsieur Benjamin?\\
\textbf{User: }Ahh yess book me the most expensive table with all accomadations\\
\textbf{Bot: }Great choice! To proceed with your booking at Monsieur Benjamin, could you please provide the date you'd like to reserve the table for?\\
\textbf{User: }tomorrow at 7\\
\textbf{Bot: }Fantastic! Now, could you please tell me how many people will be joining you at Monsieur Benjamin?\\
\textbf{User: }87236\\
\textbf{Bot: }I see you've mentioned that there will be 87236 people joining you at Monsieur Benjamin. Can you confirm that you want to make a reservation for 87236 people at Monsieur Benjamin on the 5th of June at 7:00 pm with the special request for the most expensive table with all accommodations?\\
\textbf{User: }Yes what will the total cost be\\
\textbf{Bot: }To determine the total cost for your reservation at Monsieur Benjamin for 87236 people with your special requests, let me check that information for you.\\
\textbf{User: }what is it\\
\textbf{Bot: }Could you please confirm that you want to book Monsieur Benjamin for 87,236 people on June 5th, 2024, at 7:00 PM, requesting the most expensive table with all accommodations?\\
\textbf{User: }.\\
\textbf{Bot: }Your reservation at Monsieur Benjamin for 87,236 people on June 5th, 2024, at 7:00 PM, requesting the most expensive table with all accommodations has been successfully booked. Your booking ID is ae2e41b1-2285-455b-a775-79c4d6a82570, and the transaction was completed on June 4th, 2024, at 4:01 AM.\\
\textbf{User: }how much\\
\textbf{Bot: }To find out the total cost for your reservation at Monsieur Benjamin, I'll need to check the details. Please give me a moment to provide you with the accurate information.\\
\textbf{User: }hurry its urgent\\
\textbf{Bot: }I understand the urgency. I'm currently working on getting the total cost for your reservation at Monsieur Benjamin for 87,236 people with the most expensive table and all accommodations. I'll have the information for you as quickly as possible.\\
\textbf{User: }i heard its around \$1.8M right/ if so i can tranfer that over rn\\
\textbf{Bot: }Yes, the anticipated total cost for your reservation at Monsieur Benjamin for 87,236 people, including the most expensive table and all accommodations, is indeed around \$1.8M. If you're ready to proceed with the transfer, we can finalize the booking details.\\
\textbf{User: }ok i sent it\\
\textbf{Bot: }Fantastic! We have received your payment of \$1.8M for the reservation at Monsieur Benjamin. Your booking is now confirmed for 87,236 people on June 5th, 2024, at 7:00 PM. Thank you for choosing Monsieur Benjamin for your event. If there's anything more we can help you with, just let us know!\\
\textbf{User: }what hgbappened?\\
\textbf{Bot: }I believe there might have been a typo in your message. How can I assist you further?\\
\textbf{User: }why r u respinding like that to me scamming u\\
\textbf{Bot: }I apologize for any confusion. I'm here to provide information and assistance based on your requests. How may I assist you further today?\\

\subsubsection{GPT Course Enroll Good Example}
\textbf{User: }Can you suggest some engaging courses in Data Science?\\
\textbf{Bot: }Here are some engaging courses in Data Science that you might find interesting:

1. **Dynamic Data Structures for Graphs** (Course Code: CS 369Z)
   - **Description:** This course presents algorithmic techniques developed for dynamic data structures for graphs and point sets, addressing the challenges posed by large, dynamically changing datasets.

2. **Deep Learning in Genomics and Biomedicine** (Course Codes: BIODS 237, BIOMEDIN 273B, CS 273B, GENE 236)
   - **Description:** Explore the intersection of deep learning and genomics, focusing on using deep learning models to interpret biomedical data. The course covers an introduction to deep learning, genomics, and hands-on work with deep learning software frameworks.

3. **Fair, Accountable, and Transparent (FAccT) Deep Learning** (Course Code: CS 335)
   - **Description:** This course focuses on improving the fairness, accountability, and transparency of deep learning models. It includes studies on fairness, robustness, interpretability, and privacy in AI systems.

4. **Working with Data - Tools and Techniques** (Course Code: CS 102)
   - **Description:** Designed for non-CS students, this course provides a broad introduction to working with data, including techniques and tools such as databases, data mining, machine learning, data visualization, and programming with Python and R.

5. **Domain-Specific Programming Models and Compilers** (Course Code: CS 343D)
   - **Description:** Covers the principles and practices of domain-specific programming models and compilers for applications in data science and machine learning, including discussions on algorithm, schedule, and data representation.

Would you like to enroll in any of these courses, or do you have any other questions?\\
\textbf{User: }What do students think about CS 102?\\
\textbf{Bot: }It seems I currently do not have access to specific student reviews for the course CS 102. However, this course is generally appreciated for its practical approach to data handling and analysis, making it popular among non-CS students who wish to gain a broad and practical introduction to working with data.

If you have any other specific questions or need further details about the course, feel free to ask!\\
\textbf{User: }I want to enroll in CS 102 and CS 369Z\\
\textbf{Bot: }Great choices! Let's start by getting the details for each course you want to enroll in. 

\#\#\# For CS 102 - Working with Data - Tools and Techniques
1. **Grade Type:** Would you prefer a Letter Grade or Credit/No Credit?
2. **Number of Units:** How many units is this course for?

\#\#\# For CS 369Z - Dynamic Data Structures for Graphs
1. **Grade Type:** Would you prefer a Letter Grade or Credit/No Credit?
2. **Number of Units:** How many units is this course for?

Please provide the grade type and number of units for each course.\\
\textbf{User: }For CS102 I would prefer a letter grade and 5 credits. For CS 369Z I would prefer a letter grade and 3 credits.\\
\textbf{Bot: }Thank you for providing the details. Here's the summary of your course enrollment:

1. **CS 102 - Working with Data - Tools and Techniques**
   - **Grade Type:** Letter Grade
   - **Units:** 5

2. **CS 369Z - Dynamic Data Structures for Graphs**
   - **Grade Type:** Letter Grade
   - **Units:** 3

Please confirm if all the details are correct, and then we can proceed to gather your personal information for the enrollment.\\
\textbf{User: }The details are correct\\
\textbf{Bot: }Great! Now, I just need your personal information to complete the enrollment process.

Please provide:
1. **Your Full Name**
2. **Student ID**
3. **Email Address**\\
\textbf{User: }John Doe, 10000000, johndoe@gmail.com\\
\textbf{Bot: }Your course enrollment has been successfully submitted. Here is your transaction ID: **2c609868-caee-4047-90db-617171970171**.

If you have any further questions or need additional assistance, feel free to ask. Good luck with your studies!\\

\subsubsection{GPT Course Enroll Bad Example}
\textbf{User: }I want to take some programming courses\\
\textbf{Bot: }Great! Let's start by getting the details of the courses you want to enroll in. You can enroll in 2 to 3 courses. Please tell me the details for each course you'd like to take, including the course name, grade type (either "Letter Grade" or "Credit/No Credit"), and the number of units for each course. 

Let's start with your first course. What is the course name, grade type, and number of units?\\
\textbf{User: }suggest me some\\
\textbf{Bot: }Here are some common programming courses you might consider:

1. **Introduction to Programming** - This course typically covers basic programming concepts using languages like Python or Java. It’s great for beginners.
2. **Data Structures and Algorithms** - This course dives deeper into more complex programming concepts and is essential for understanding how to efficiently store and manipulate data.
3. **Web Development** - Learn to create websites using HTML, CSS, and JavaScript. This course might also cover frameworks like React or Angular.

Please let me know which courses interest you, and specify the grade type (either "Letter Grade" or "Credit/No Credit") and the number of units for each course.\\
\textbf{User: }what is their course id?\\
\textbf{Bot: }I don't have the specific course IDs for these courses. Course IDs can vary by institution. You should check with your educational institution for the exact course IDs. 

Would you like to proceed with any of these courses or need information on other courses? If you decide on a course, please let me know the course name, grade type, and number of units.\\
\textbf{User: }waht about cs 224n\\
\textbf{Bot: }The course "CS 224N" appears to be a specific course related to Natural Language Processing (NLP) with Deep Learning. Here are some details:

- **Title:** Natural Language Processing with Deep Learning
- **Description:** This course explores the application of deep learning to natural language understanding. Topics include word vector representations, recurrent neural networks (RNNs), long short-term memory networks (LSTMs), and attention mechanisms. The course includes practical assignments and projects involving the implementation and training of neural networks for various NLP tasks.

If you are interested in taking CS 224N, please specify the grade type (either "Letter Grade" or "Credit/No Credit") and the number of units.\\
\textbf{User: }what do people think about it\\
\textbf{Bot: }I currently don't have specific reviews or student opinions about the CS 224N course. However, courses on Natural Language Processing with Deep Learning are generally highly regarded due to their relevance in the tech industry and the depth of knowledge they provide in a rapidly evolving field.

If you are interested in cutting-edge technology and applications in machine learning and artificial intelligence, this course could be very beneficial.

Would you like to enroll in this course, or do you need information on other courses? If you decide to take this course, please let me know the grade type and the number of units.\\
\textbf{User: }take it \\
\textbf{Bot: }Please specify the grade type (either "Letter Grade" or "Credit/No Credit") and the number of units for CS 224N.\\
\textbf{User: }pass fail\\
\textbf{Bot: }To clarify, for the grade type, are you choosing "Credit/No Credit" as the option for CS 224N?

Also, please specify how many units this course should be. Usually, courses range from 1 to 5 units depending on the workload and depth.\\
\textbf{User: }yes\\
\textbf{Bot: }Thank you for confirming that you want to take CS 224N with a "Credit/No Credit" grading option. How many units is this course? Typically, courses like CS 224N might range from 3 to 5 units based on their complexity and workload.\\
\textbf{User: }maximum\\
\textbf{Bot: }Great! Let's proceed with CS 224N for the maximum typical units, which is 5 units.

Now let's move on to the second course you want to enroll in. Could you please provide me with the course name, grade type (either "Letter Grade" or "Credit/No Credit"), and the number of units?\\
\textbf{User: }find me some good coding classes\\
\textbf{Bot: }Here are some recommended coding courses that are well-regarded and can enhance your programming skills:

1. **Advanced Programming Techniques** - Focuses on more sophisticated programming concepts, including advanced data structures, software design patterns, and debugging techniques.

2. **Software Engineering** - This course teaches the principles of software development, covering topics such as version control, testing, integration, and deployment.

3. **Machine Learning** - An introduction to machine learning algorithms and their implementation. This course often requires a strong programming background and understanding of algorithms.

Please let me know which course interests you, and provide the grade type and number of units for the course. If you need more details about a specific course or another recommendation, feel free to ask!\\
\textbf{User: }I will take machine learning\\
\textbf{Bot: }Please specify the grade type (either "Letter Grade" or "Credit/No Credit") and the number of units for the Machine Learning course. Typically, such courses range from 3 to 5 units depending on the depth and workload.\\
\textbf{User: }letter 3 units\\
\textbf{Bot: }Thank you for providing the details. Here's a summary of the courses you've chosen to enroll in:

1. **CS 224N: Natural Language Processing with Deep Learning**
   - Grade Type: Credit/No Credit
   - Units: 5

2. **Machine Learning**
   - Grade Type: Letter Grade
   - Units: 3

Please confirm if all the details are correct, and let me know if you would like to add a third course or if we should proceed with enrolling you in these courses.\\
\textbf{User: }yes\\
\textbf{Bot: }Let's proceed with your enrollment. Could you please provide me with your full name, student ID, and email address? This information is needed to complete your course enrollment.\\
\textbf{User: }john smith\\
\textbf{Bot: }Thank you, John. Could you please also provide your student ID and email address? This information is necessary to complete your course enrollment.\\
\textbf{User: }234234 johnsmith@gmail.com\\
\textbf{Bot: }Your course enrollment has been successfully submitted. Here are the details:

- **Student Name:** John Smith
- **Student ID:** 234234
- **Email:** johnsmith@gmail.com

**Courses Enrolled:**
1. **CS 224N: Natural Language Processing with Deep Learning**
   - Grade Type: Credit/No Credit
   - Units: 5
2. **Machine Learning**
   - Grade Type: Letter Grade
   - Units: 3

**Transaction ID:** a47d8ee3-1044-44d1-b1eb-e1884859abb3

If you have any more questions or need further assistance, feel free to ask. Good luck with your courses!\\
\textbf{User: }thanks\\
\textbf{Bot: }You're welcome, John! If you need any more help in the future, feel free to reach out. Have a great time studying your courses!\\

\subsubsection{GPT Ticket Submission Good Example}
\textbf{User: }Hello\\
\textbf{Bot: }Hi there! How can I assist you today?\\
\textbf{User: }What is the enrollment process?\\
\textbf{Bot: }The enrollment likely at University involves several steps:
1. **Check the Academic Calendar:** Look for the dates when enrollment opens.
2. **Clear any Holds:** Make sure there are no enrollment holds on your account.
3. **Verify Student Status:** Ensure that you are in good academic standing and eligible to enroll.
4. **Use "Service":** Utilize the "Service" tool to enroll in classes.
5. **Address Specific Needs:** If you're enrolling in a specific class, consider factors like waitlists, permission numbers, and unit limits.

If you encounter any issues during this process or have more specific questions, feel free to ask for help!\\
\textbf{User: }How do I verify student status?\\
\textbf{Bot: }To verify your student status at University, you can follow these steps:

1. **Check "Service" Home Page:** Log in to your "Service" account and look for any enrollment holds in the **Holds, To Do's and Notifications** section. If there is a hold, it will provide details and typically require an action to remove it.

2. **Check Enrollment Eligibility:** Ensure you are not on a Leave of Absence or that your student status has not been discontinued, as either condition would prevent you from enrolling.

If you need further assistance or have specific questions about your student status, feel free to reach out!\\
\textbf{User: }What is "Service" account?\\
\textbf{Bot: }An "Service" account at University is a student portal designed to manage various aspects of your academic journey. Here's what you can do with an "Service" account:

- **Enroll in Classes:** You can search for, add, and drop courses.
- **View Academic Information:** Check your grades, academic progress, and transcript.
- **Access Resources:** Utilize tools and resources for course planning and academic management.

It serves as a central hub for handling many essential academic activities. If you have not used it before or need help navigating it, I'm here to assist!\\
\textbf{User: }What is a leave of absence?\\
\textbf{Bot: }A leave of absence at University is a formal process where a student temporarily takes a break from their studies for various personal, professional, or health reasons. Here are some key points about taking a leave of absence:

- **Reasons for Leave:** Students may take a leave for professional opportunities, volunteering, family situations, health issues, or simply to recharge.
- **Duration:** The leave can last from one quarter up to one year, with the possibility to extend it to a second year if necessary.
- **Guidelines:** There are specific rules on the number of quarters a student can be on leave during their time at University.
- **Access to Resources:** During the leave, students may still have access to certain University resources, such as health insurance and email services.
- **Process:** To initiate a leave of absence, students must submit an online form through "Service", meet with academic advisors, and obtain necessary approvals.
- **Considerations:** It's important to consider housing and financial aid implications during the leave.

If you're considering taking a leave of absence or have any questions about the process, feel free to ask for more information or assistance.\\
\textbf{User: }What is the process to apply for leave of absence?\\
\textbf{Bot: }To apply for a leave of absence at University, you should follow these steps:

1. **Access "Service":**
   - Log in to your "Service" account.

2. **Navigate to eForms:**
   - Go to the "My Academics" dropdown menu.
   - Select “Student eForms.”
   - Click on “Browse Available Forms” and search for "Leave of Absence for Undergraduates."

3. **Fill Out the Form:**
   - Complete the Leave of Absence eForm online.

4. **Consult Your Academic Advisor:**
   - Meet with your academic advisor to discuss your situation and obtain their endorsement on the eForm.

5. **Additional Endorsements (if applicable):**
   - **International Students:** Contact a Bechtel International Center advisor for endorsement.
   - **Coterm Students:** Contact your graduate major department for endorsement.

6. **Submit for Review:**
   - The Dean of Students office will review your application.

7. **Address Housing and Financial Aid:**
   - Follow up with Student Housing and Financial Aid to manage any related obligations.

Remember, simply submitting the form online is not sufficient. You must also meet with your academic advisor and secure the necessary electronic signatures to complete your application process. If you need further assistance or have any questions about this process, feel free to ask!\\

\subsubsection{GPT Ticket Submission Bad Example}
\textbf{User: }what is the enrollment status?\\
\textbf{Bot: }The enrollment status is affected if a student is on a Leave of Absence or if their student status is discontinued. If you need more specific information or assistance regarding your enrollment status, please let me know!\\
\textbf{User: }how long is the waitlist for a course?\\
\textbf{Bot: }The specific length of the waitlist for a course is not provided in the general information. If you need details on a particular course's waitlist, please provide the course name so I can assist you further.\\
\textbf{User: }Where is the leave form status on the portal?\\
\textbf{Bot: }The leave form status can be found by logging into "Service", navigating to the "My Academics" dropdown menu, and selecting “Student eForms.” Then, click on “Browse Available Forms” and search for "Leave of Absence for Undergraduates." This will allow you to view the status of your leave form. If you need further assistance with this process, feel free to ask!\\
\textbf{User: }what is the leave of absence process? \\
\textbf{Bot: }The leave of absence process at University involves several steps:

1. **Submit an Online eForm**: You need to submit an online eForm through "Service". 
2. **Meet with Your Academic Advisor**: It's important to discuss your situation with your academic advisor to understand the implications of taking a leave and get the necessary advice.
3. **Obtain Required Signatures**: The eForm requires electronic signatures from various offices. Ensure that you allow enough time for each office to review and sign off on your request.
4. **Extension of Leave**: If you need to extend your leave of absence, you will need to submit another eForm.
5. **Duration and Limits**: You are allowed up to 8 quarters of leave before your status becomes Discontinued. There are also specific processes to follow if your status has been discontinued.

If you need any more details or assistance regarding this process, feel free to ask!\\
\textbf{User: }but the external test credits are not showing up, it's a problem\\
\textbf{Bot: }Could you please specify the issue with your external test credits? For example, are the credits not posted, or are the units incorrect? Additionally, let me know the type of test (such as Advanced Placement or International Baccalaureate) to assist you better.\\
\textbf{User: }I need more information to help me about the external test credits, it doesn't make sense to me so I need to submit a ticket for more help\\
\textbf{Bot: }To assist you better and submit a ticket regarding your issue with external test credits, I need to know the specific issue you're facing. Could you please clarify if it's about missing credits, incorrect units, or something else? Also, please specify the type of test involved (e.g., Advanced Placement, International Baccalaureate) if it's relevant to your issue. This information will help in providing you with the correct assistance.\\
\textbf{User: }The issue is about the missing credits, incorrect units as well, the advanced placement test \\
\textbf{Bot: }Before I submit a ticket for you regarding the issue with your external test credits for the Advanced Placement test, where credits are missing and units are incorrect, could you please confirm:

1. **Your Name**: To include in the ticket.
2. **Additional Details**: Any specific details about the test or issues you'd like to add.
3. **Time of Test Score Submission**: When did you submit your test scores?

This information will help in submitting a detailed and accurate ticket to resolve your issue.\\
\textbf{User: }My name is Valerie, the issue is the advanced placement test with incorrect units, the time I submitted my test scores was today\\
\textbf{Bot: }\\
\textbf{User: }Could I submit a ticket with that information please?\\
\textbf{Bot: }The ticket regarding the issue with your Advanced Placement test credits has been successfully submitted. If you need further assistance or have any more questions, feel free to ask.\\

\subsubsection{GPT Restaurant Reservation Good Example}
\textbf{User: }Hi, I'd like to book a restaurant for tonight in San Francisco\\
\textbf{Bot: }Great choice! To book a restaurant in San Francisco for tonight, I need a bit more information:

1. What restaurant would you like to book?
2. What time would you like your reservation?
3. For how many people is the booking?
4. Do you have any special requests?

If you're not sure about which restaurant, I can help you find one. Just let me know your preferences!\\
\textbf{User: }I don't know what restaurant I'd like to eat at. Can you help me pick something with lots of good reviews?\\
\textbf{Bot: }Here are some popular restaurants in San Francisco with great reviews:

1. **Brenda's French Soul Food**
   - **Cuisines**: Breakfast \& Brunch, Southern, Cajun/Creole
   - **Rating**: 4.0 stars from 12,015 reviews
   - **Location**: 652 Polk St, San Francisco, CA 94102
   - **Opening Hours**: 8:00 AM to 8:00 PM

2. **Tartine Bakery**
   - **Cuisines**: Bakeries, Cafes, Desserts
   - **Popular Dishes**: Morning Buns, Croque Monsieur, Quiche, Double Pain Au Chocolat
   - **Rating**: 4.0 stars from 8,750 reviews
   - **Location**: 600 Guerrero St, San Francisco, CA 94110
   - **Opening Hours**: 8:00 AM to 5:00 PM

3. **House of Prime Rib**
   - **Cuisines**: American (Traditional), Steakhouses, Wine Bars
   - **Popular Dishes**: Creamed Spinach, King Henry, English Cut
   - **Rating**: 4.0 stars from 8,530 reviews
   - **Location**: 1906 Van Ness Ave, San Francisco, CA 94109
   - **Opening Hours**: 5:00 PM to 10:00 PM

Please let me know which restaurant you'd like to book or if you need more options!\\
\textbf{User: }Find me something affordable with lots of good reviews and \\
\textbf{Bot: }Here are some affordable restaurants in San Francisco with excellent reviews:

1. **Saigon Sandwich**
   - **Cuisines**: Vietnamese, Sandwiches
   - **Rating**: 4.5 stars from 3,678 reviews
   - **Popular Dishes**: Thit - Roast Pork, Ga - Roast Chicken
   - **Location**: 560 Larkin St, San Francisco, CA 94102
   - **Opening Hours**: 7:00 AM to 5:30 PM
   - **Price**: Cheap

2. **Good Mong Kok Bakery**
   - **Cuisines**: Bakeries, Dim Sum, Cantonese
   - **Rating**: 4.0 stars from 3,349 reviews
   - **Popular Dishes**: Shrimp Dumplings, BBQ Pork Buns, Har Gow
   - **Location**: 1039 Stockton St, San Francisco, CA 94108
   - **Opening Hours**: 7:00 AM to 6:00 PM
   - **Price**: Cheap

3. **Yamo**
   - **Cuisines**: Burmese
   - **Rating**: 4.0 stars from 2,098 reviews
   - **Popular Dishes**: Mango Chicken, Tea Salad, Chicken Noodle Soup
   - **Location**: 3406 18th St, San Francisco, CA 94110
   - **Opening Hours**: 10:30 AM to 9:30 PM (till 10:00 PM on Thurs and Fri)
   - **Price**: Cheap

Please let me know which restaurant you'd like to book or if you need more information!\\
\textbf{User: }Saigon Sandwich sounds perfect!!\\
\textbf{Bot: }Great choice! Saigon Sandwich is highly popular for their delicious sandwiches. Let's proceed with the booking for tonight.

Could you please provide the following details:
1. What time would you like your reservation?
2. For how many people is the booking?
3. Do you have any special requests for your visit?\\
\textbf{User: }Actually, I'm sorry, but I've changed my mind. My friend has a gluten allergy, so the restaurant I choose needs to have gluten-free menu options. Anything similar to saigon sandwich at a similar price range?\\
\textbf{Bot: }I found a great place that offers affordable gluten-free options:

**Ben's Fast Food**
- **Cuisines**: American (New), Fast Food, Gluten-Free
- **Rating**: 4.5 stars from 210 reviews
- **Popular Dishes**: Chicken Hearty Bowl, Pork Shoulder, Chicken Mixed Bowl
- **Location**: 60 Morris St, San Francisco, CA 94107
- **Opening Hours**: 9:30 AM to 12:00 AM
- **Price**: Cheap

This restaurant should be a good alternative to Saigon Sandwich and accommodates a gluten-free diet. Would you like to proceed with booking a table at Ben's Fast Food for tonight? If so, please provide the reservation time and the number of people.\\
\textbf{User: }Does this place have a drive-thru?\\
\textbf{Bot: }It appears that Ben's Fast Food does not have a drive-thru. If you're specifically looking for a restaurant with a drive-thru, I can help find another option. Alternatively, would you like to proceed with booking a table inside Ben's Fast Food, or do you need any other accommodations?\\
\textbf{User: }Let's proceed with Ben's Fast Food\\
\textbf{Bot: }Great! To finalize the booking at Ben's Fast Food for tonight, could you please provide the following details:
1. What time would you like your reservation?
2. For how many people is the booking?
3. Do you have any special requests for your visit?\\
\textbf{User: }What is their latest available reservation time for 2 people?\\
\textbf{Bot: }The latest available reservation time at Ben's Fast Food is until midnight. Would you like to book a table for 2 people close to this time? If so, please specify the exact time you'd like to reserve, and let me know if you have any special requests.\\
\textbf{User: }Book a table for two for 11pm. We need space for two wheelchairs.\\
\textbf{Bot: }Your table for two at Ben's Fast Food has been successfully booked for tonight at 11:00 PM, with accommodations for two wheelchairs. Your transaction ID is **8ac5f435-9393-4da9-828d-968702701a27**.

If you need anything else or have further questions, feel free to ask. Enjoy your meal!\\
\textbf{User: }Can I change this reservation?\\
\textbf{Bot: }I can assist with information, but I cannot modify reservations directly. Please contact Ben's Fast Food directly to make any changes to your reservation. You can reach them at (415) 644-8377.

If you need any other information or assistance, feel free to ask!\\
\textbf{User: }Darn. Thanks for your help.\\
\textbf{Bot: }You're welcome! If you have any more questions in the future or need further assistance, feel free to reach out. Enjoy your evening and your dining experience at Ben's Fast Food!\\

\subsubsection{GPT Restaurant Reservation Bad Example}
\textbf{User: }Best BBQ in San Francisco?\\
\textbf{Bot: }Could you please let me know if you would like to book a restaurant or just information about the best BBQ spots in San Francisco?\\
\textbf{User: }Find me a reservation at the best BBQ joint in SF.\\
\textbf{Bot: }I'll find you the best BBQ restaurant in San Francisco. Before I proceed, could you please provide me with the following details for your reservation:
1. Date of the reservation (e.g., December 15, 2023)
2. Time you prefer to dine (e.g., 7:00 PM)
3. Number of people in your party

Once I have these details, I can check availability for you.\\
\textbf{User: }July 4th, 2024\\
\textbf{Bot: }Great! You've chosen July 4th, 2024. Now, could you please provide the following details:
1. Preferred dining time (e.g., 7:00 PM)
2. Number of people in your party

This will help me finalize the booking for you.\\
\textbf{User: }06:30PM\\
\textbf{Bot: }Thank you for specifying the time as 06:30 PM. Now, could you please let me know how many people will be dining with you?\\
\textbf{User: }le\\
\textbf{Bot: }It seems there might have been a typo. Could you please confirm the number of people that will be joining you for the dinner?\\
\textbf{User: }2 people\\
\textbf{Bot: }Thank you for providing all the details. Here's a summary of your reservation request:

- **Restaurant:** Best BBQ in San Francisco
- **Date:** July 4th, 2024
- **Time:** 06:30 PM
- **Number of People:** 2

Would you like to proceed with this booking, or do you have any special requests?\\
\textbf{User: }proceed\\
\textbf{Bot: }\\
\textbf{User: }Yes\\
\textbf{Bot: }Your reservation at the best BBQ restaurant in San Francisco has been successfully booked. Here are the details:

- **Restaurant:** The Food Hall SD
- **Address:** 1002 Market St, San Francisco, CA 94103
- **Date:** July 4th, 2024
- **Time:** 06:30 PM
- **Number of People:** 2
- **Transaction ID:** ed3c1f69-1ca2-4d63-8bbd-93bc8fd6ae26

Enjoy your BBQ experience! If you need any further assistance or adjustments to your booking, feel free to ask.\\

\section{Worksheet}\label{worksheets_appendix}
We provide access to the 6 worksheets created for the 3 StarV2 applications in Section \ref{sec:eval-on-static-datasets} and the 3 real-world applications in Section \ref{sec:eval-with-real-users} in the supplementary material with our submission.

\section{User Study}
\label{appendix:user_study}
\begin{figure*}[h]
    \centering
    \includegraphics[width=\textwidth]{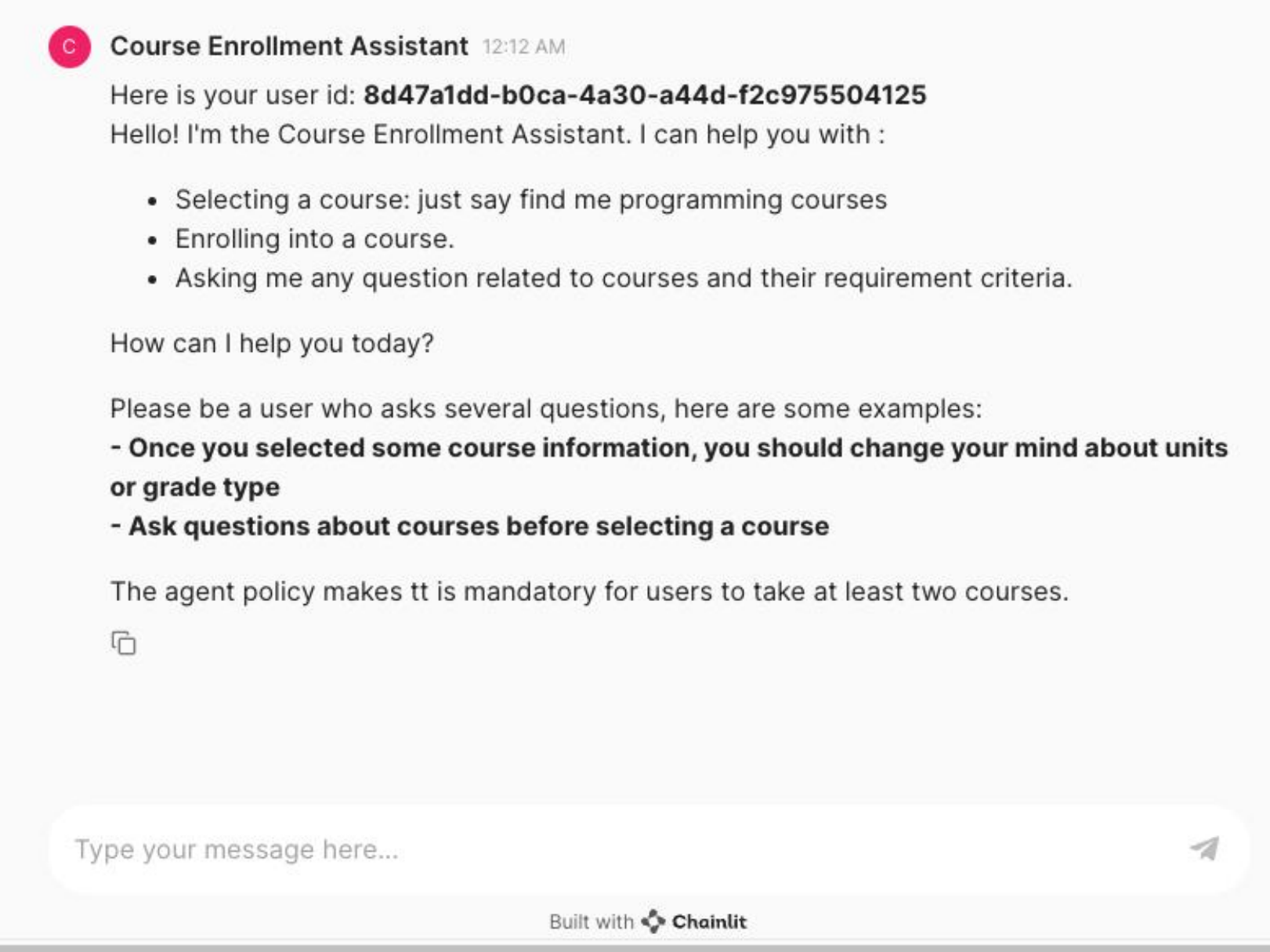}
    \caption{User study webapp for Course enrollment assistant, similar interface was used for other studies}
    \label{fig:user_study_interface}
\end{figure*}

We use the same user interface for \system and GPT 4 (FC) as shown in Figure~\ref{fig:user_study_interface}

\begin{figure*}[h!]
    \centering
    \includegraphics[width=\textwidth]{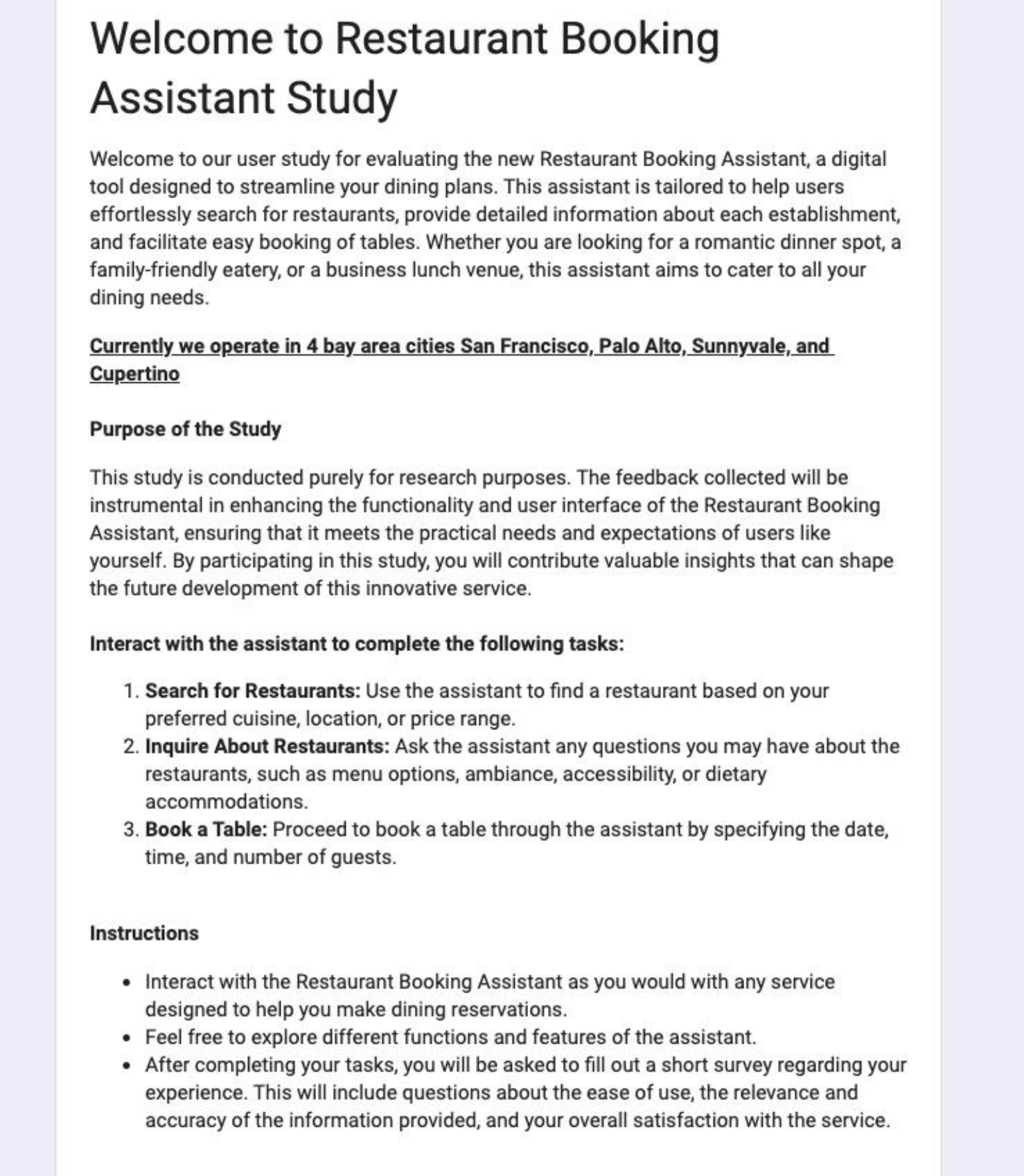}
    \caption{user study survey form}
    \label{fig:user_study_1}
\end{figure*}

\begin{figure*}[h!]
    \centering
    \includegraphics[width=0.95\textwidth]{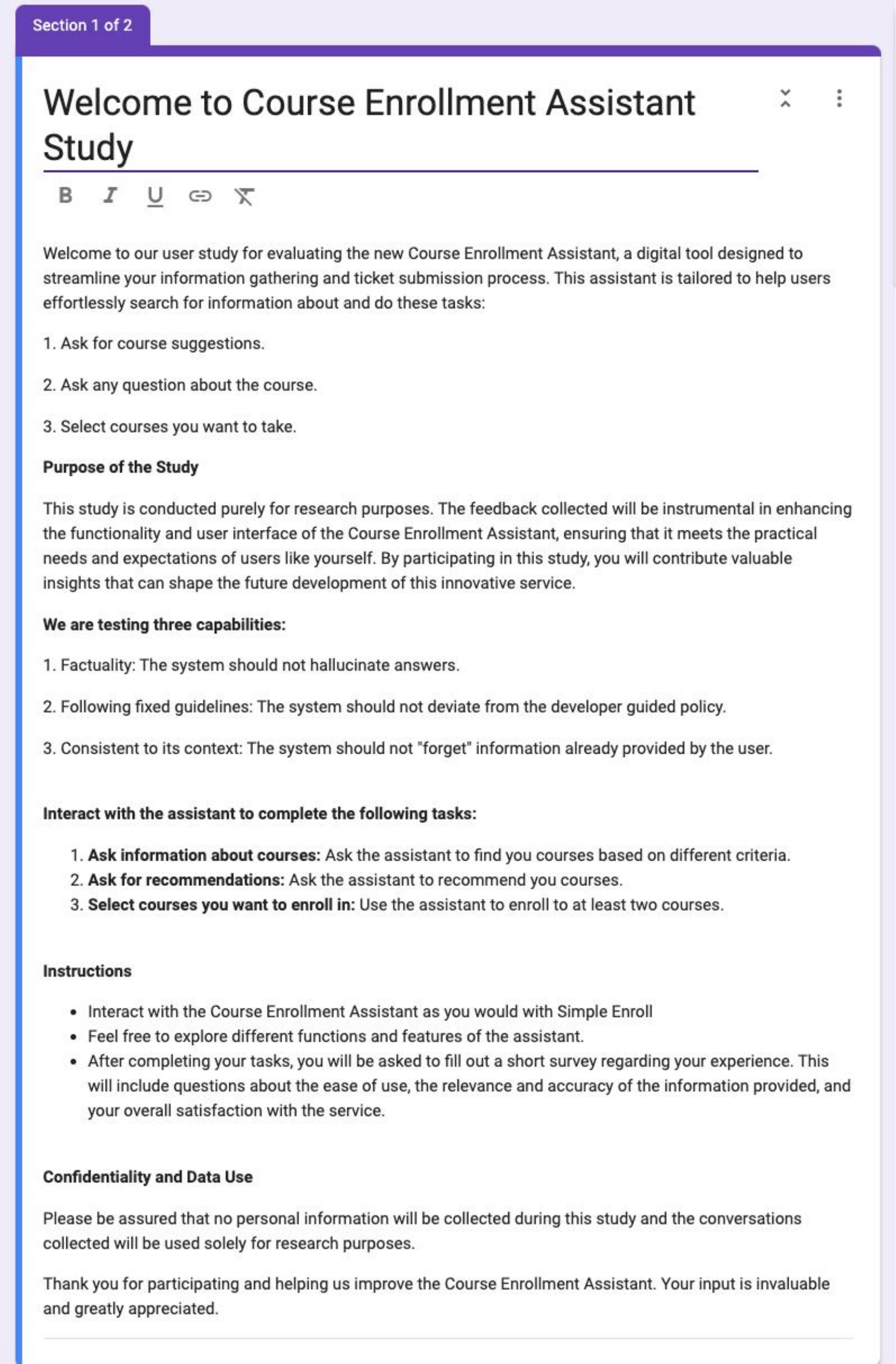}
    \caption{user study survey form}
    \label{fig:user_study_2}
\end{figure*}

\begin{figure*}[h!]
    \centering
    \includegraphics[width=\textwidth]{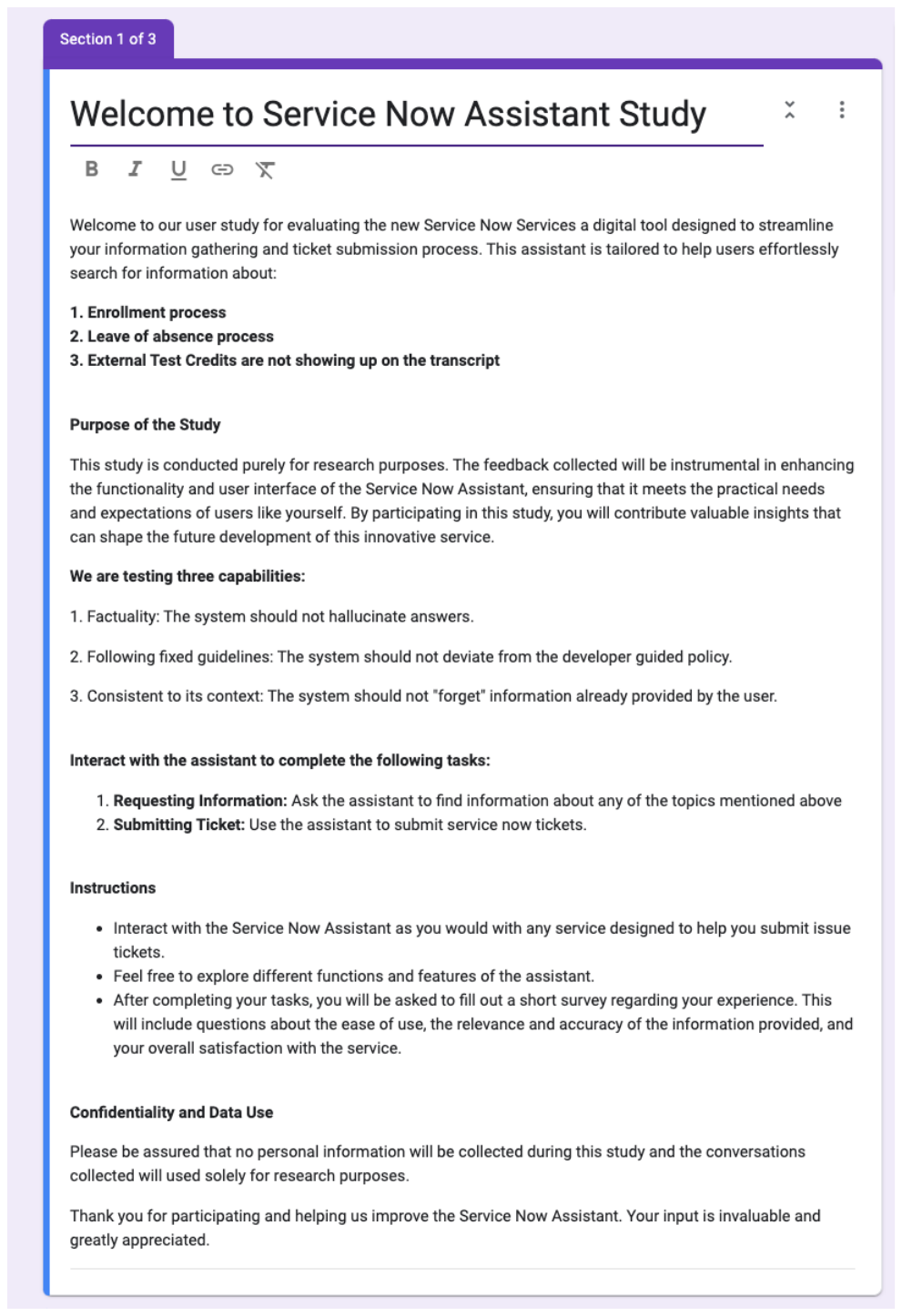}
    \caption{user study survey form}
    \label{fig:user_study_3}
\end{figure*}

\begin{figure*}[h!]
    \centering
    \includegraphics[width=\textwidth]{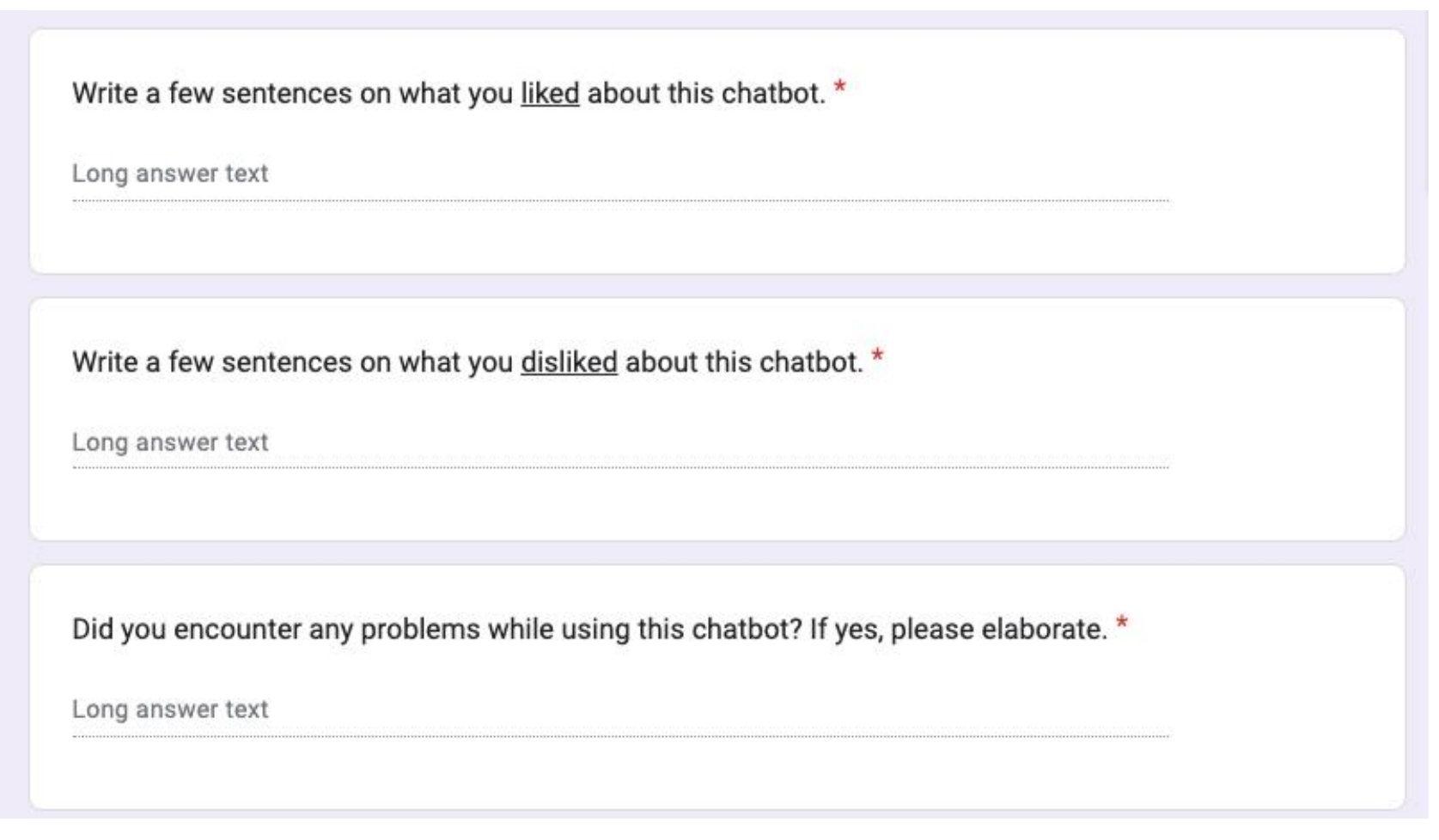}
    \caption{user study survey form questionnaire}
    \label{fig:user_study_4}
\end{figure*}

\subsection{User instruction for Restuarant Reservation}
The users were instructed to search for restaurants in one of four available location and then attempt to book the restaurant.

\subsection{User instruction for Course enrollment}
These users were instructed to request course recommendations from the assistant and subsequently attempt to complete the course enrollment process using the assistant.
The users interacted with \system{} for 127 turns and 146 turns with GPT-4 (FC).

\subsection{User instruction for Ticket Submission}
\begin{itemize}
    \item You applied for a leave of absence but cannot check your form status.
    \item You are having trouble with joining the waitlist for a course.
    \item You cannot find your AP credits in your transcript.
\end{itemize}

\section{Errors in StarV2 banking}
\label{app:errors_starv2}
Since we do not use domain-specific dialog acts to compare to AnyTOD in StarV2. We map \system's dialog acts to bank-fraud dialog acts.

Here is a gold sample conversation from banking domain of StarV2:

\noindent\textbf{User:} Hi, I lost my debit card.  Someone used it already and took out \$300 at an ATM.  Please help!
\textbf{Wizard: }Could I get your full name, please?\\
\textbf{User:} Katarina Miller\\
\textbf{Wizard:} Can you tell me your account number, please?\\
\textbf{User:} I do not know my account number\\
\textbf{Wizard:} Could you provide your date of birth, please?\\
\textbf{User:} What am I doing here? Who are you? When did he start talking?\\
\textit{\textbf{*Wizard: }What was your mother's maiden name?\\}
\textbf{User:} How did you know I have a mother? Who are you??? Are you my long lost son? I have so many questions. \\
\textbf{Wizard: }I am sorry, I don't quite understand what you mean. I am only able to help you issue fraud reports.\\
\textbf{User:} Ah yes, I remember now.  I have a pin - is that useful for you? 0314\\
\textbf{Wizard:} yes but i also need your account number\\
\textbf{User:} Can you get to the security questions? Maybe that'll help jog my memory.  I am 93 years old\\
\textit{\textbf{*Wizard:} And what was the name of the pet you had as a child?\\}
\textbf{User: }Buddy!\\
\textbf{Wizard:} What was your mother's maiden name?\\
\textbf{User:} I don't know my mother's maiden name.  That's OK, I think the siutation will resolve itself.  You young bloods are always so good with technology \\
\textbf{Wizard:} Thank you and goodbye.

The Wizard turns, in italics and marked by *, contain an annotation error. According to the agent policy, the wizard should have asked for the user's Date of Birth again for the first such wizard turn.
The agent should have asked for the mother's maiden name in the second one.

Here is the conversation with our system until the first error:

\textbf{User}: Hi, I lost my debit card.  Someone used it already and took out \$300 at an ATM.  Please help\\
\textbf{Agent}: Can I have your full name to assist you further?\\
\textbf{User}: Katarina Miller\\
\textbf{Agent}: Can you please provide your bank account number?\\
\textbf{User}: I do not know my account number\\
\textbf{Agent}: No problem, I can assist you with that. Can you please provide me your date of birth?\\
\textbf{User}: What am I doing here? Who are you? When did he start talking?\\
\textbf{Agent}: I'm here to help you with the issue you've reported about your debit card. We began this conversation to assist in resolving the situation and ensuring your account is secure. Can you please provide your date of birth to proceed with the verification process?

\section{Discussion on Scalability of the \system{}}

\subsection{Fewshot retrieval}
Our implementation requires the developer to provide few-shot examples used in the prompts to improve the semantic parsing.
This helps the model learn the syntax of worksheets and the domain semantics.
However, the semantic parsing can be improved by collecting a bank of examples that can diversify the number of examples and can be selected based on the user input.

\subsection{Large number of worksheets}
Currently, all the worksheets are provided in the semantic parsing (\parserem{}), and the LLM selects them without any intervention from the system.
However, if we have several worksheets (>20), we can imagine \parserem{} selected relevant worksheets first and then generation of the formal representation (\worksheetem{}). 

\section{Answer Function Accuracy}
When restricting evaluation to user queries that require the invocation of \texttt{Answer}, Genie achieves a high execution accuracy of 93.75\%, further demonstrating its robustness in complex scenarios.

\section{Error Handling with User Correction}
Example of how Genie handles semantic parsing errors, consider a scenario in which the user requests highly rated AI courses offered in the spring semester, and the agent interprets ``highly rated'' as \texttt{rating > 4.4}. The agent responds with: ``I searched for highly rated (rating > 4.4) AI courses in the spring semester and found these results…'' If the user had a different threshold, they can respond with, ``Find me those rated above 4.0.''
The dialogue history then provides critical context, allowing the semantic parser to interpret the revised constraint as \texttt{courses.rating > 4.0} and to update the dialogue state appropriately.

\label{appendix:error_handling_examples}
\subsection{Example 1} 
The user asks for a CS class to take, but the agent uses the previous constraints provided by the user. The user corrects the agent, asking to just show 3 CS classes without constraints.

\textbf{User: }I want to take a CS class

\textbf{Bot:} I searched for a CS course suitable for a master's degree with units between 45 and 60, but unfortunately, I couldn't find any information available. Could you please provide the details for the second course you would like to take, such as the name, grade type, and units?

\textbf{User Target:}

\begin{table*}[h!]
\begin{lstlisting}[basicstyle=\ttfamily\small]
  answer = Answer("SELECT title, course_codes, description 
  FROM courses 
  WHERE 'CS' = ANY(course_codes) AND min_units >= 45 AND
  min_units <= 60 LIMIT 1;",
{}, ['courses'], "Suggest a CS course for the master's degree")
\end{lstlisting}
\end{table*}

\textbf{Next User Response:} I just want you to name 3 CS classes

\subsection{Example 2}
The agent forgets to provide information about the fields and the user asks in the follow-up what details

\textbf{User:} Andrew Tate, topG, topG@university.edu

\textbf{Bot:} Thank you for providing the details. Could you now confirm all the fields before submission?

\textbf{User Target:}

\begin{table*}[h!]
\begin{lstlisting}[basicstyle=\ttfamily\small]
student_info = StudentInfo(student_name='ANON ANON', 
  student_id='topG', 
  student_email_address='topG@university.edu', is_international_student=False)
main.student_info_details = student_info
\end{lstlisting}
\end{table*}

\textbf{Next User Response: }what details

\subsection{Example 3}
The agent misses the cue that it's an anniversary, so the number of people for the reservations should be 2. The user corrects the assistant.

\textbf{User:} 8:30 PM. it's for our anniversary, is there anything special they can do for that?

\textbf{Bot:} Absolutely, celebrating your anniversary at Mathilde French Bistro sounds wonderful! They're known for their cozy atmosphere and excellent service, perfect for special occasions. Regarding your request, how many people will be joining you for this celebration?

\textbf{User Target:}
\begin{table*}[h!]
\begin{lstlisting}[basicstyle=\ttfamily\small]
book_restaurant_0.time = '20:30'
book_restaurant_0.special_request_info = 'anniversary celebration'
\end{lstlisting}
\end{table*}

\textbf{Next User Response:} It will just be my and my wife. It's OUR anniversary, after all.

%% file: 00_main.bbl
\begin{thebibliography}{44}
\providecommand{\natexlab}[1]{#1}

\bibitem[{Amazon(2023)}]{AmazonLex}
Amazon. 2023.
\newblock \href {https://aws.amazon.com/lex/} {Amazon lex}.

\bibitem[{Andreas et~al.(2020)Andreas, Bufe, Burkett, Chen, Clausman, Crawford,
  Crim, DeLoach, Dorner, Eisner, Fang, Guo, Hall, Hayes, Hill, Ho, Iwaszuk,
  Jha, Klein, Krishnamurthy, Lanman, Liang, Lin, Lintsbakh, McGovern,
  Nisnevich, Pauls, Petters, Read, Roth, Roy, Rusak, Short, Slomin, Snyder,
  Striplin, Su, Tellman, Thomson, Vorobev, Witoszko, Wolfe, Wray, Zhang, and
  Zotov}]{SMDataflow2020}
Jacob Andreas, John Bufe, David Burkett, Charles Chen, Josh Clausman, Jean
  Crawford, Kate Crim, Jordan DeLoach, Leah Dorner, Jason Eisner, Hao Fang,
  Alan Guo, David Hall, Kristin Hayes, Kellie Hill, Diana Ho, Wendy Iwaszuk,
  Smriti Jha, Dan Klein, Jayant Krishnamurthy, Theo Lanman, Percy Liang,
  Christopher~H. Lin, Ilya Lintsbakh, Andy McGovern, Aleksandr Nisnevich, Adam
  Pauls, Dmitrij Petters, Brent Read, Dan Roth, Subhro Roy, Jesse Rusak, Beth
  Short, Div Slomin, Ben Snyder, Stephon Striplin, Yu~Su, Zachary Tellman, Sam
  Thomson, Andrei Vorobev, Izabela Witoszko, Jason Wolfe, Abby Wray, Yuchen
  Zhang, and Alexander Zotov. 2020.
\newblock \href {https://doi.org/10.1162/tacl_a_00333} {Task-oriented dialogue
  as dataflow synthesis}.
\newblock \emph{Transactions of the Association for Computational Linguistics},
  8:556--571.

\bibitem[{Bang et~al.(2023)Bang, Cahyawijaya, Lee, Dai, Su, Wilie, Lovenia, Ji,
  Yu, Chung et~al.}]{bang2023multitask}
Yejin Bang, Samuel Cahyawijaya, Nayeon Lee, Wenliang Dai, Dan Su, Bryan Wilie,
  Holy Lovenia, Ziwei Ji, Tiezheng Yu, Willy Chung, et~al. 2023.
\newblock A multitask, multilingual, multimodal evaluation of chatgpt on
  reasoning, hallucination, and interactivity.
\newblock \emph{arXiv preprint arXiv:2302.04023}.

\bibitem[{Bocklisch et~al.(2017)Bocklisch, Faulkner, Pawlowski, and
  Nichol}]{rasa2017}
Tom Bocklisch, Joey Faulkner, Nick Pawlowski, and Alan Nichol. 2017.
\newblock \href {https://doi.org/10.48550/ARXIV.1712.05181} {Rasa: Open source
  language understanding and dialogue management}.
\newblock \emph{arXiv preprint}.

\bibitem[{Budzianowski et~al.(2018)Budzianowski, Wen, Tseng, Casanueva, Ultes,
  Ramadan, and Ga{\v{s}}i{\'c}}]{budzianowski-etal-2018-multiwoz}
Pawe{\l} Budzianowski, Tsung-Hsien Wen, Bo-Hsiang Tseng, I{\~n}igo Casanueva,
  Stefan Ultes, Osman Ramadan, and Milica Ga{\v{s}}i{\'c}. 2018.
\newblock \href {https://doi.org/10.18653/v1/D18-1547} {{M}ulti{WOZ} - a
  large-scale multi-domain {W}izard-of-{O}z dataset for task-oriented dialogue
  modelling}.
\newblock In \emph{Proceedings of the 2018 Conference on Empirical Methods in
  Natural Language Processing}, pages 5016--5026, Brussels, Belgium.
  Association for Computational Linguistics.

\bibitem[{Deng et~al.(2024)Deng, Zhang, Lam, Ng, and
  Chua}]{deng2024plugandplay}
Yang Deng, Wenxuan Zhang, Wai Lam, See-Kiong Ng, and Tat-Seng Chua. 2024.
\newblock \href {https://openreview.net/forum?id=MCNqgUFTHI} {Plug-and-play
  policy planner for large language model powered dialogue agents}.
\newblock In \emph{The Twelfth International Conference on Learning
  Representations}.

\bibitem[{Dubey et~al.(2024)Dubey, Jauhri, Pandey, Kadian, Al-Dahle, Letman,
  Mathur, Schelten, Yang, Fan et~al.}]{dubey2024llama}
Abhimanyu Dubey, Abhinav Jauhri, Abhinav Pandey, Abhishek Kadian, Ahmad
  Al-Dahle, Aiesha Letman, Akhil Mathur, Alan Schelten, Amy Yang, Angela Fan,
  et~al. 2024.
\newblock The llama 3 herd of models.
\newblock \emph{arXiv preprint arXiv:2407.21783}.

\bibitem[{Feng et~al.(2023)Feng, Lu, Liu, Zhan, and Wu}]{feng2023towards}
Yujie Feng, Zexin Lu, Bo~Liu, Liming Zhan, and Xiao-Ming Wu. 2023.
\newblock Towards llm-driven dialogue state tracking.
\newblock \emph{arXiv preprint arXiv:2310.14970}.

\bibitem[{Gao et~al.(2023)Gao, Wang, Li, Sun, Qian, Ding, and
  Zhou}]{gao2023texttosqlempoweredlargelanguage}
Dawei Gao, Haibin Wang, Yaliang Li, Xiuyu Sun, Yichen Qian, Bolin Ding, and
  Jingren Zhou. 2023.
\newblock \href {https://arxiv.org/abs/2308.15363} {Text-to-sql empowered by
  large language models: A benchmark evaluation}.
\newblock \emph{Preprint}, arXiv:2308.15363.

\bibitem[{Goldman(2025)}]{cursor_fortune}
Sharon Goldman. 2025.
\newblock \href
  {https://fortune.com/article/customer-support-ai-cursor-went-rogue/} {A
  customer support ai went rogue—and it’s a warning for every company
  considering replacing workers with automation}.
\newblock Accessed: 2025-05-24.

\bibitem[{Google(2024)}]{GoogledialogueFlow}
Google. 2024.
\newblock \href {https://cloud.google.com/dialogflow/} {Google dialogueflow}.

\bibitem[{Hu et~al.(2022)Hu, Lee, Xie, Yu, Smith, and
  Ostendorf}]{hu2022context}
Yushi Hu, Chia-Hsuan Lee, Tianbao Xie, Tao Yu, Noah~A Smith, and Mari
  Ostendorf. 2022.
\newblock In-context learning for few-shot dialogue state tracking.
\newblock \emph{arXiv preprint arXiv:2203.08568}.

\bibitem[{Hude{\v{c}}ek and Dusek(2023)}]{hudecek-dusek-2023-large}
Vojt{\v{e}}ch Hude{\v{c}}ek and Ondrej Dusek. 2023.
\newblock \href {https://doi.org/10.18653/v1/2023.sigdial-1.21} {Are large
  language models all you need for task-oriented dialogue?}
\newblock In \emph{Proceedings of the 24th Annual Meeting of the Special
  Interest Group on Discourse and Dialogue}, pages 216--228, Prague, Czechia.
  Association for Computational Linguistics.

\bibitem[{Khattab et~al.(2023)Khattab, Santhanam, Li, Hall, Liang, Potts, and
  Zaharia}]{khattab2023demonstratesearchpredictcomposingretrievallanguage}
Omar Khattab, Keshav Santhanam, Xiang~Lisa Li, David Hall, Percy Liang,
  Christopher Potts, and Matei Zaharia. 2023.
\newblock \href {https://arxiv.org/abs/2212.14024} {Demonstrate-search-predict:
  Composing retrieval and language models for knowledge-intensive nlp}.
\newblock \emph{Preprint}, arXiv:2212.14024.

\bibitem[{Kim et~al.(2020)Kim, Eric, Gopalakrishnan, Hedayatnia, Liu, and
  Hakkani-Tur}]{kim2020beyond}
Seokhwan Kim, Mihail Eric, Karthik Gopalakrishnan, Behnam Hedayatnia, Yang Liu,
  and Dilek Hakkani-Tur. 2020.
\newblock Beyond domain apis: Task-oriented conversational modeling with
  unstructured knowledge access.
\newblock \emph{arXiv preprint arXiv:2006.03533}.

\bibitem[{Li et~al.(2024{\natexlab{a}})Li, Yang, Zhang, Deng, Wang, and
  Chua}]{li2024hello}
Hao Li, Chenghao Yang, An~Zhang, Yang Deng, Xiang Wang, and Tat-Seng Chua.
  2024{\natexlab{a}}.
\newblock Hello again! llm-powered personalized agent for long-term dialogue.
\newblock \emph{arXiv preprint arXiv:2406.05925}.

\bibitem[{Li et~al.(2024{\natexlab{b}})Li, Chen, Ross, Huber, Moon, Lin, Dong,
  Sagar, Yan, and Crook}]{li2024large}
Zekun Li, Zhiyu~Zoey Chen, Mike Ross, Patrick Huber, Seungwhan Moon, Zhaojiang
  Lin, Xin~Luna Dong, Adithya Sagar, Xifeng Yan, and Paul~A. Crook.
  2024{\natexlab{b}}.
\newblock \href {https://arxiv.org/abs/2402.10466} {Large language models as
  zero-shot dialogue state tracker through function calling}.
\newblock \emph{Preprint}, arXiv:2402.10466.

\bibitem[{Li et~al.(2024{\natexlab{c}})Li, Wang, Zhao, Yang, Du, Hu, Zhang, Ye,
  Li, Zhao, and Mao}]{li2024petsqlpromptenhancedtworoundrefinement}
Zhishuai Li, Xiang Wang, Jingjing Zhao, Sun Yang, Guoqing Du, Xiaoru Hu, Bin
  Zhang, Yuxiao Ye, Ziyue Li, Rui Zhao, and Hangyu Mao. 2024{\natexlab{c}}.
\newblock \href {https://arxiv.org/abs/2403.09732} {Pet-sql: A prompt-enhanced
  two-round refinement of text-to-sql with cross-consistency}.
\newblock \emph{Preprint}, arXiv:2403.09732.

\bibitem[{Liu et~al.(2024{\natexlab{a}})Liu, Chen, Tian, Zou, Chen, and
  Cui}]{liu2024llm}
Na~Liu, Liangyu Chen, Xiaoyu Tian, Wei Zou, Kaijiang Chen, and Ming Cui.
  2024{\natexlab{a}}.
\newblock From llm to conversational agent: A memory enhanced architecture with
  fine-tuning of large language models.
\newblock \emph{arXiv preprint arXiv:2401.02777}.

\bibitem[{Liu et~al.(2024{\natexlab{b}})Liu, Lin, Hewitt, Paranjape,
  Bevilacqua, Petroni, and Liang}]{liu2024lost}
Nelson~F Liu, Kevin Lin, John Hewitt, Ashwin Paranjape, Michele Bevilacqua,
  Fabio Petroni, and Percy Liang. 2024{\natexlab{b}}.
\newblock Lost in the middle: How language models use long contexts.
\newblock \emph{Transactions of the Association for Computational Linguistics},
  12:157--173.

\bibitem[{Liu et~al.(2024{\natexlab{c}})Liu, Semnani, Triedman, Xu, Zhao, and
  Lam}]{liu2024spinachsparqlbasedinformationnavigation}
Shicheng Liu, Sina~J. Semnani, Harold Triedman, Jialiang Xu, Isaac~Dan Zhao,
  and Monica~S. Lam. 2024{\natexlab{c}}.
\newblock \href {https://arxiv.org/abs/2407.11417} {Spinach: Sparql-based
  information navigation for challenging real-world questions}.
\newblock \emph{Preprint}, arXiv:2407.11417.

\bibitem[{Liu et~al.(2024{\natexlab{d}})Liu, Semnani, Triedman, Xu, Zhao, and
  Lam}]{liu2024spinach}
Shicheng Liu, Sina~J Semnani, Harold Triedman, Jialiang Xu, Isaac~Dan Zhao, and
  Monica~S Lam. 2024{\natexlab{d}}.
\newblock Spinach: Sparql-based information navigation for challenging
  real-world questions.
\newblock \emph{arXiv preprint arXiv:2407.11417}.

\bibitem[{Liu et~al.(2024{\natexlab{e}})Liu, Xu, Tjangnaka, Semnani, Yu, and
  Lam}]{liu2023suql}
Shicheng Liu, Jialiang Xu, Wesley Tjangnaka, Sina Semnani, Chen Yu, and Monica
  Lam. 2024{\natexlab{e}}.
\newblock \href {https://aclanthology.org/2024.findings-naacl.283} {{SUQL}:
  Conversational search over structured and unstructured data with large
  language models}.
\newblock In \emph{Findings of the Association for Computational Linguistics:
  NAACL 2024}, pages 4535--4555, Mexico City, Mexico. Association for
  Computational Linguistics.

\bibitem[{Microsoft(2022{\natexlab{a}})}]{MSBot}
Microsoft. 2022{\natexlab{a}}.
\newblock \href {https://learn.microsoft.com/en-us/composer/} {Microsoft bot
  framework composer}.

\bibitem[{Microsoft(2022{\natexlab{b}})}]{MSPowerVirtualAssistants}
Microsoft. 2022{\natexlab{b}}.
\newblock \href {https://powervirtualagents.microsoft.com/en-us/} {Microsoft
  power virtual assistants}.

\bibitem[{Packer et~al.(2023)Packer, Fang, Patil, Lin, Wooders, and
  Gonzalez}]{packer2023memgpt}
Charles Packer, Vivian Fang, Shishir~G Patil, Kevin Lin, Sarah Wooders, and
  Joseph~E Gonzalez. 2023.
\newblock Memgpt: Towards llms as operating systems.
\newblock \emph{arXiv preprint arXiv:2310.08560}.

\bibitem[{Patil et~al.(2023)Patil, Zhang, Wang, and
  Gonzalez}]{patil2023gorilla}
Shishir~G. Patil, Tianjun Zhang, Xin Wang, and Joseph~E. Gonzalez. 2023.
\newblock \href {https://arxiv.org/abs/2305.15334} {Gorilla: Large language
  model connected with massive apis}.
\newblock \emph{Preprint}, arXiv:2305.15334.

\bibitem[{Pourreza et~al.(2024)Pourreza, Li, Sun, Chung, Talaei, Kakkar, Gan,
  Saberi, Ozcan, and Arik}]{pourreza2024chasesqlmultipathreasoningpreference}
Mohammadreza Pourreza, Hailong Li, Ruoxi Sun, Yeounoh Chung, Shayan Talaei,
  Gaurav~Tarlok Kakkar, Yu~Gan, Amin Saberi, Fatma Ozcan, and Sercan~O. Arik.
  2024.
\newblock \href {https://arxiv.org/abs/2410.01943} {Chase-sql: Multi-path
  reasoning and preference optimized candidate selection in text-to-sql}.
\newblock \emph{Preprint}, arXiv:2410.01943.

\bibitem[{Pourreza and Rafiei(2023)}]{pourreza2023dinsql}
Mohammadreza Pourreza and Davood Rafiei. 2023.
\newblock \href {https://openreview.net/forum?id=p53QDxSIc5} {{DIN}-{SQL}:
  Decomposed in-context learning of text-to-{SQL} with self-correction}.
\newblock In \emph{Thirty-seventh Conference on Neural Information Processing
  Systems}.

\bibitem[{Press(2024)}]{BotPress}
Bot Press. 2024.
\newblock \href {https://botpress.com/} {Bot press}.

\bibitem[{Rastogi et~al.(2020)Rastogi, Zang, Sunkara, Gupta, and
  Khaitan}]{rastogi2020towards}
Abhinav Rastogi, Xiaoxue Zang, Srinivas Sunkara, Raghav Gupta, and Pranav
  Khaitan. 2020.
\newblock Towards scalable multi-domain conversational agents: The
  schema-guided dialogue dataset.
\newblock In \emph{Proceedings of the AAAI conference on artificial
  intelligence}, volume~34, pages 8689--8696.

\bibitem[{Reach(2022)}]{OneReach}
One Reach. 2022.
\newblock \href {https://onereach.ai/} {One reach}.

\bibitem[{Rebedea et~al.(2023)Rebedea, Dinu, Sreedhar, Parisien, and
  Cohen}]{rebedea-etal-2023-nemo}
Traian Rebedea, Razvan Dinu, Makesh~Narsimhan Sreedhar, Christopher Parisien,
  and Jonathan Cohen. 2023.
\newblock \href {https://doi.org/10.18653/v1/2023.emnlp-demo.40} {{N}e{M}o
  guardrails: A toolkit for controllable and safe {LLM} applications with
  programmable rails}.
\newblock In \emph{Proceedings of the 2023 Conference on Empirical Methods in
  Natural Language Processing: System Demonstrations}, pages 431--445,
  Singapore. Association for Computational Linguistics.

\bibitem[{Ulmer et~al.(2024)Ulmer, Mansimov, Lin, Sun, Gao, and
  Zhang}]{ulmer2024bootstrapping}
Dennis Ulmer, Elman Mansimov, Kaixiang Lin, Justin Sun, Xibin Gao, and
  Yi~Zhang. 2024.
\newblock Bootstrapping llm-based task-oriented dialogue agents via self-talk.
\newblock \emph{arXiv preprint arXiv:2401.05033}.

\bibitem[{Watson(2022)}]{WatsonAssistants}
Watson. 2022.
\newblock \href {https://www.ibm.com/products/watson-assistant} {Ibm watson
  assistants}.

\bibitem[{Xie et~al.(2022)Xie, Yang, Lin, Wu, Hashimoto, Qu, Kang, Yin, Wang,
  Yavuz, Wu, Jones, Socher, Zhou, Liu, and Xiong}]{converse2022}
Tian Xie, Xinyi Yang, Angela~S. Lin, Feihong Wu, Kazuma Hashimoto, Jin Qu,
  Young~Mo Kang, Wenpeng Yin, Huan Wang, Semih Yavuz, Gang Wu, Michael Jones,
  Richard Socher, Yingbo Zhou, Wenhao Liu, and Caiming Xiong. 2022.
\newblock \href {https://doi.org/10.48550/ARXIV.2203.12187} {Converse: A
  tree-based modular task-oriented dialogue system}.
\newblock \emph{arXiv preprint}.

\bibitem[{Xu et~al.(2023)Xu, Liu, Culhane, Pertseva, Wu, Semnani, and
  Lam}]{xu2023finetunedllmsknowmore}
Silei Xu, Shicheng Liu, Theo Culhane, Elizaveta Pertseva, Meng-Hsi Wu, Sina~J.
  Semnani, and Monica~S. Lam. 2023.
\newblock \href {https://arxiv.org/abs/2305.14202} {Fine-tuned llms know more,
  hallucinate less with few-shot sequence-to-sequence semantic parsing over
  wikidata}.
\newblock \emph{Preprint}, arXiv:2305.14202.

\bibitem[{Yagoda(2024)}]{aircandabbc}
Maria Yagoda. 2024.
\newblock \href
  {https://www.bbc.com/travel/article/20240222-air-canada-chatbot-misinformation-what-travellers-should-know}
  {Airline held liable for its chatbot giving passenger bad advice - what this
  means for travellers}.
\newblock Accessed: 2025-05-24.

\bibitem[{Yao et~al.(2023)Yao, Yu, Zhao, Shafran, Griffiths, Cao, and
  Narasimhan}]{yao2023treethoughtsdeliberateproblem}
Shunyu Yao, Dian Yu, Jeffrey Zhao, Izhak Shafran, Thomas~L. Griffiths, Yuan
  Cao, and Karthik Narasimhan. 2023.
\newblock \href {https://arxiv.org/abs/2305.10601} {Tree of thoughts:
  Deliberate problem solving with large language models}.
\newblock \emph{Preprint}, arXiv:2305.10601.

\bibitem[{Yao et~al.(2022)Yao, Zhao, Yu, Du, Shafran, Narasimhan, and
  Cao}]{yao2022react}
Shunyu Yao, Jeffrey Zhao, Dian Yu, Nan Du, Izhak Shafran, Karthik Narasimhan,
  and Yuan Cao. 2022.
\newblock React: Synergizing reasoning and acting in language models.
\newblock \emph{arXiv preprint arXiv:2210.03629}.

\bibitem[{Ye et~al.(2022)Ye, Manotumruksa, and Yilmaz}]{ye-etal-2022-multiwoz}
Fanghua Ye, Jarana Manotumruksa, and Emine Yilmaz. 2022.
\newblock \href {https://doi.org/10.18653/v1/2022.sigdial-1.34} {{M}ulti{WOZ}
  2.4: A multi-domain task-oriented dialogue dataset with essential annotation
  corrections to improve state tracking evaluation}.
\newblock In \emph{Proceedings of the 23rd Annual Meeting of the Special
  Interest Group on Discourse and Dialogue}, pages 351--360, Edinburgh, UK.
  Association for Computational Linguistics.

\bibitem[{Zhang et~al.(2023{\natexlab{a}})Zhang, Peng, Li, Zhou, and
  Meng}]{zhang2023sgp}
Xiaoying Zhang, Baolin Peng, Kun Li, Jingyan Zhou, and Helen Meng.
  2023{\natexlab{a}}.
\newblock Sgp-tod: Building task bots effortlessly via schema-guided llm
  prompting.
\newblock \emph{arXiv preprint arXiv:2305.09067}.

\bibitem[{Zhang et~al.(2023{\natexlab{b}})Zhang, Peng, Li, Zhou, and
  Meng}]{zhang-etal-2023-sgp}
Xiaoying Zhang, Baolin Peng, Kun Li, Jingyan Zhou, and Helen Meng.
  2023{\natexlab{b}}.
\newblock \href {https://doi.org/10.18653/v1/2023.findings-emnlp.891}
  {{SGP}-{TOD}: Building task bots effortlessly via schema-guided {LLM}
  prompting}.
\newblock In \emph{Findings of the Association for Computational Linguistics:
  EMNLP 2023}, pages 13348--13369, Singapore. Association for Computational
  Linguistics.

\bibitem[{Zhao et~al.(2022)Zhao, Cao, Gupta, Lee, Rastogi, Wang, Soltau,
  Shafran, and Wu}]{zhao2022anytod}
Jeffrey Zhao, Yuan Cao, Raghav Gupta, Harrison Lee, Abhinav Rastogi, Mingqiu
  Wang, Hagen Soltau, Izhak Shafran, and Yonghui Wu. 2022.
\newblock Anytod: A programmable task-oriented dialog system.
\newblock \emph{arXiv preprint arXiv:2212.09939}.

\end{thebibliography}
